\documentclass[final,5p,times,twocolumn]{elsarticle}

\usepackage{hyperref}
\usepackage{mathtools,multirow,soul,comment,afterpage}
\usepackage{amsmath,amssymb,amsthm,amsfonts}
\usepackage{booktabs,textcomp}
\usepackage{graphicx,wrapfig,float,stfloats,subcaption}
\usepackage{algorithm,algpseudocode}
\usepackage[table]{xcolor}

\journal{Journal of Information Applications and Security}

\begin{document}

\begin{frontmatter}

\title{FRIDA: Free-Rider Detection using Privacy Attacks}

\author[inst1]{Pol G. Recasens}
\author[inst2]{Ádám Horváth}
\author[inst1]{Alberto Gutierrez-Torre}
\author[inst1,inst3]{Jordi Torres}
\author[inst3,inst1]{Josep Ll. Berral}
\author[inst2,inst4]{Balázs Pejó}

\affiliation[inst1]{organization={Barcelona Supercomputing Center},
            addressline={Plaça d'Eusebi Güell, 1-3},
            city={Barcelona},
            postcode={08034},
            country={Spain}}

\affiliation[inst2]{organization={Laboratory of Cryptography and System Security, Dept. of Networked Systems and Services, Faculty of Electrical Engineering and Informatics, Budapest University of Technology and Economics},
            addressline={Műegyetem rkp. 3},
            city={Budapest},
            postcode={1111},
            country={Hungary}}

\affiliation[inst3]{organization={Universitat Politècnica de Catalunya},
            addressline={Carrer de Jordi Girona, 1-3},
            city={Barcelona},
            postcode={08034},
            country={Spain}}

\affiliation[inst4]{organization={HUN-REN Hungarian Research Network, Office for Supported Research Groups, HUN-REN-BME Information Systems Research Group},
            addressline={Piarista utca 4},
            city={Budapest},
            postcode={1052},
            country={Hungary}}

\begin{abstract}
Federated learning is increasingly popular as it enables multiple parties with limited datasets and resources to train a machine learning model collaboratively. However, similar to other collaborative systems, federated learning is vulnerable to free-riders — participants who benefit from the global model without contributing. Free-riders compromise the integrity of the learning process and slow down the convergence of the global model, resulting in increased costs for honest participants. To address this challenge, we propose FRIDA: \textbf{f}ree-\textbf{ri}der \textbf{d}etection using privacy \textbf{a}ttacks. Instead of focusing on implicit effects of free-riding, FRIDA utilizes membership and property inference attacks to directly infer evidence of genuine client training. Our extensive evaluation demonstrates that FRIDA is effective across a wide range of scenarios.
\end{abstract}

\begin{keyword}
Federated Learning \sep Free-Riding \sep Privacy Attacks \sep Membership Inference \sep Property Inference
\end{keyword}

\end{frontmatter}

\section{Introduction}
\label{sec:intro}

Our daily lives are closely connected to computers, as we rely on them for both leisure and work. This broad use has enabled companies to collect vast amounts of user data at low cost, driving remarkable advancements in machine learning (ML)~\cite{zhou2021machine}. To leverage diverse datasets without centralizing them, federated learning (FL)~\cite{mcmahan2017communication} allows multiple data holders to collaboratively train an ML model, enhancing prediction performance while preserving data privacy.

\begin{figure}[!tb]
    \centering
    \includegraphics[width=0.35\textwidth]{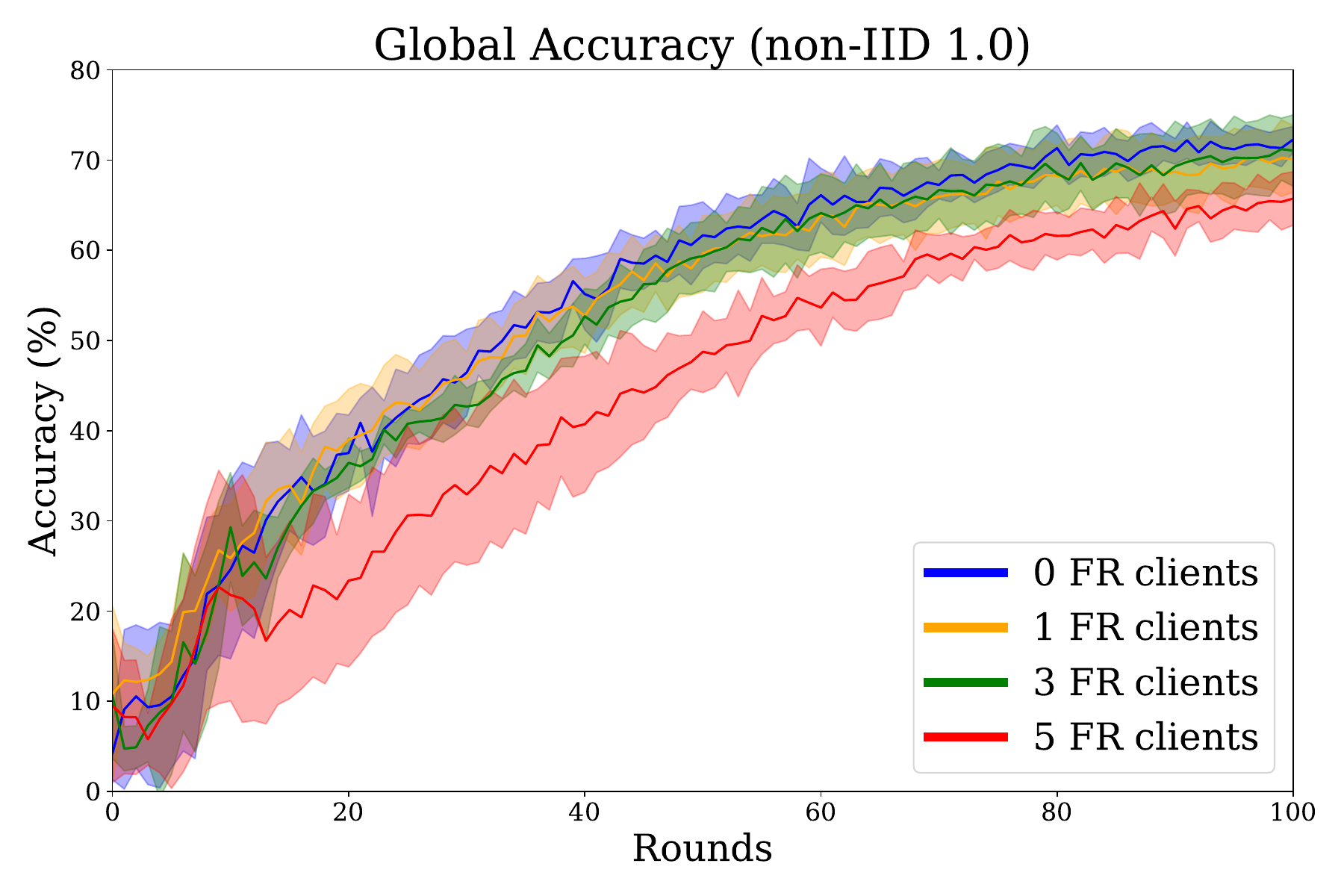}
    \caption{Convergence of the global model in a cross-silo FL scenario with 10 clients and 0/1/3/5 FR clients. The model corresponds to AlexNet, trained on CIFAR-10 distributed non-IID among clients.} 
    \label{fig:delay}
\end{figure}

However, the collaborative nature of FL introduces the problem of free-riding (FR)~\cite{hardin2003free,krishnan2004impact}. In FL, free-riders~\cite{lin2019free,fraboni2021free,zhu2021advanced} are participants who conceal their lack of contribution during training while still benefiting from the final aggregated global model. Although their objective is not to disrupt the learning process, their behavior slows down model convergence (see Figure~\ref{fig:delay}) and undermines fairness, forcing honest participants to invest more resources in terms of time and computation. This issue is particularly critical in domains such as healthcare, where data is scarce and highly sensitive. Here, free-riders can exploit access to models trained on valuable private datasets without contributing anything in return.

The challenge of detecting free-riders is deeply intertwined with several other fundamental aspects of FL, namely contribution evaluation, security, privacy, and data distribution~\cite{kairouz2021advances}.

In terms of contribution evaluation (CE), free-rider detection can be viewed as an extreme case of quantifying each participant’s value~\cite{huang2020exploratory}. Although the final model in FL is shared, externalities can still arise (e.g., by selling it), making fair reward distribution non-trivial. The gold standard for this is the Shapley Value~\cite{winter2002shapley,rozemberczki2022shapley}, which is computationally intractable. While approximations like Data Shapley~\cite{ghorbani2019data} and Distributional Shapley~\cite{ghorbani2020distributional} exist, they often focus on ranking the best and worst contributors rather than precisely identifying zero-contribution participants.

From a security perspective, free-riding is a subtle threat compared to aggressive poisoning~\cite{tolpegin2020data} and backdoor attacks~\cite{bagdasaryan2020backdoor}, where malicious clients actively inject unintended behavior into models. While free-riders do not aim to corrupt the model, they still compromise the system's integrity and fairness. Defenses against malicious attacks often involve Byzantine Fault Tolerant (BFT) aggregation schemes like KRUM~\cite{blanchard2017machine}, FoolsGold~\cite{fung2018mitigating}, and Bulyan~\cite{guerraoui2018hidden}, which are designed for overt attacks and may not be suited for detecting passive free-riding.

Regarding privacy, FL also faces inherent risks, as shared model updates can leak sensitive information via membership inference attacks (MIA)~\cite{shokri2017membership}, property inference attacks (PIA)~\cite{parisot2021property}, and reconstruction attacks (Rec)~\cite{zhu2019deep}. To mitigate these risks, techniques like secure aggregation (SA)~\cite{bonawitz2017practical} and differential privacy (DP)~\cite{pejo2022guide} are employed. SA uses secure multiparty computation to hide individual gradients, while DP adds noise to obscure individual contributions. However, these privacy-enhancing technologies create a dilemma: by obscuring individual updates, they also make it significantly harder to perform contribution evaluation or detect attackers such as free-riders, who can hide behind the curtain of privacy.

Regarding data distribution, compounding all these issues is data heterogeneity. Participants' datasets are often non-homogeneous (non-IID), which can reduce the effectiveness of standard aggregation methods like FedAvg~\cite{mcmahan2017communication}. This heterogeneity makes it difficult to distinguish a free-rider from an honest participant whose local data simply diverges from the global distribution. While alternative techniques like FedProx~\cite{li2020federated} and Scaffold~\cite{karimireddy2020scaffold} address heterogeneity, robustly handling it remains an open challenge that complicates attack detection and contribution analysis. In essence, such statistical diversity could provide natural camouflage for free-riders, making their detection significantly more difficult.

While these FL challenges are often addressed independently, they significantly influence one another. For instance, many BFT aggregation techniques and CE mechanisms are incompatible with privacy-protection methods like SA, as SA conceals the individual gradients these schemes rely on. Our objective is to bridge the security and privacy perspectives to address FR detection. Rather than relying on the indirect effects of free-riding (e.g., update statistics), we propose leveraging privacy attacks to directly verify whether a client has genuinely participated in model training.

\subsubsection*{Contribution}

In this paper, we address the relatively understudied security problem of free-riding in FL by proposing four detection mechanisms based on privacy attacks. These mechanisms identify clients who fail to contribute to the model's training by directly inferring information indicative of real training. The proposed methods include \textit{loss-based} and \textit{cosine-based} approaches, which rely on membership inference, as well as \textit{consistency-based} and \textit{diversity-based} approaches, which rely on property inference. Our contributions are detailed below.

\begin{itemize}
    \item We propose four methods that leverage privacy attacks to detect FR clients. Unlike prior work, our approach targets directly the root cause \textemdash lack of genuine training contributions \textemdash instead of relying on indirect effects of free-riding strategies.
    % \item We demonstrate that FRIDA is agnostic to specific FR strategies by employing three different FR techniques in our empirical experiments.
    % \item We show that FRIDA can detect advanced FR strategies that remain undetected by feature-based detection mechanisms in the IID scenario.
    \item We demonstrate that FRIDA effectively detects advanced FR strategies that remain undetected by feature-based mechanisms, particularly in the IID scenario.
    \item We establish a novel connection between FL security and privacy by applying privacy attacks to address a security problem.  
\end{itemize}

We validate our approach through extensive experiments across multiple datasets and model architectures. We show that FRIDA remains compatible with privacy-preserving techniques such as local DP, and under selfish adaptive attackers who adapt to detection mechanisms. 

% We show that FRIDA is compatible with privacy-preserving architectures, such as a two-server implementation that decouples client identities from their updates. Additionally, we perform an ablation study to assess how applying differential privacy impacts FRIDA's detection performance.

\subsubsection*{Paper Organization}

In Section~\ref{sec:bg-rw}, we present the related work and background.  
In Section~\ref{sec:mod}, we detail how privacy attacks can be adapted to detect free-riders. 
In Section~\ref{sec:exp}, we describe the experimental setup and present the corresponding results.
Finally, in Section~\ref{sec:con}, we conclude our work, highlight its limitations, and propose several future directions. 

\section{Background \& Related Work}
\label{sec:bg-rw}

In this section we review the background and related work, and summarise the core techniques used in our paper, with an emphasis on misbehaviour detection and privacy attacks.

\subsection{Misbehaviour Detection}

Defences in FL broadly fall into two families: mitigation and detection. Mitigation modifies the aggregation technique~\cite{guerraoui2024byzantine}, typically using robust statistics such as the median~\cite{pillutla2022robust}, or comparing local updates either to each other~\cite{fung2018mitigating} or to a ground truth/baseline~\cite{xie2019zeno} such as the global model. These comparisons generally rely on distance/similarity metrics, including $\ell_2$ distance~\cite{blanchard2017machine} or cosine similarity~\cite{guerraoui2018hidden}.

Detection methods, in contrast, estimate a per-client \emph{quality} (or suspicion) score and then modify the client sampling by filtering or down-weighting clients via relative or absolute thresholds~\cite{wan2022shielding}. Many existing approaches (e.g., FLTrust~\cite{cao2020fltrust}) rely on indirect characteristics of client updates, and thus often miss less-deviated adversaries, including ACE~\cite{xu2024ace} and free-riders, whose updates remain close to the benign updates.

\subsubsection{Free-riding}

The authors in~\cite{fraboni2021free} provided a theoretical framework that studies how the global model converges to the optimal in the presence of free-riders, albeit at a slower convergence rate. ~\cite{wang2022assessing} distinguishes between two groups of FR clients, named anonymous and selfish. Anonymous free-riders do not have computational power or data, while selfish free-riders have resources and avoid using them. The primary cause for anonymous FR is a lack of resources, whereas selfish FR mainly occurs due to economic incentives. In this paper we focus on anonymous FR, named FR in the rest of the paper, which appear in critical scenarios involving sensitive data.

In a nutshell, FR strategies \cite{lin2019free, zhu2021advanced, fraboni2021free} mainly rely on adding perturbations to the global parameters or to the model updates, aiming to blend with the stochastic behavior of the training process. The authors in~\cite{lin2019free} introduced the \textit{random weights} attack, where the gradient update is generated by randomly sampling each parameter uniformly. They also proposed the \textit{advanced delta weights} attack, which constructs the gradient update by adding noise to the aggregated gradient. A similar attack was presented in~\cite{zhu2021advanced}, where geometrical properties in IID scenarios are used to scale the update accordingly. Finally, the authors in~\cite{fraboni2021free} introduced the \textit{disguised free-rider}, where noise is added to the previous global model parameters to mimic the SGD noise of honest clients. 

In this paper, we consider the three FR strategies developed in~\cite{lin2019free},~\cite{fraboni2021free}, and~\cite{zhu2021advanced}, denoted as $\mathcal{F}^L$, $\mathcal{F}^F$, and $\mathcal{F}^Z$, which have been theoretically shown to fall within the stochasticity of an honest client update. More details about these attacks are provided in ~\ref{app:attacks}.

% While the first two attacks are relatively simple, adding plain noise to the gradient or the model, they have been theoretically shown to fall within the stochasticity of an honest client update, making their detection considerably more challenging than that of malicious clients. Furthermore, our addresses the core issue of free-riding by identifying the absence of genuine training, regardless of the specific FR strategy employed. Therefore, we argue that FRIDA is agnostic to the FR strategy.

\subsubsection{Free-rider Detection}

Compared to misbehaviour detection, the literature on free-rider detection remains relatively limited, with most works framing the problem as outlier detection. Lin et al.~\cite{lin2019free} introduced the Deep Autoencoding Gaussian Mixture Model (DAGMM) as a detection mechanism, which was later used in~\cite{huang2022delta,wang2023frad}. DAGMM, however, requires training an autoencoder on each set of client updates, and its memory demand grows with both the model size and the number of clients, making it computationally challenging beyond small networks. We analyze this scalability bottleneck in Section~\ref{sec:comp_overhead} and report STD-DAGMM baselines in smaller-scale experiments in Ablation~\ref{ablation:num_clients}. Free-rider detection has also been explored with feature-based detection \cite{zhu2021advanced}, which focuses on indirect effects of free-riders, such as the $\ell_2$ norm or standard deviation.

Other lines of work detect free-riders via log-based auditing of client activity~\cite{nguyen2024stake} or through reward-allocation mechanisms that distribute models proportionally to contributions~\cite{lyu2020collaborative,xu2020reputation,wang2023pass}. These schemes operate on information beyond raw model updates (e.g., detailed logs, reward accounting), and are therefore orthogonal to update-based detection.  

Among hybrid approaches, FRAD~\cite{wang2023frad} combines DAGMM with a contribution-evaluation (CE) method, which helps improve detection when the FR proportion is high; however, the detection mechanism remains DAGMM. In this work, we focus on update-based detection, and thus use STD-DAGMM and feature-based detection as the directly comparable baselines. Log-based and reward-based signals are complementary and can be integrated with FRIDA in future designs.

\begin{table}[t]
\footnotesize
\centering
\caption{Free-rider detection families. Update-based methods rely only on per-round client updates; log/incentive methods require additional information.}
\label{tab:fr-comparison}
\begin{tabular}{l l}
\toprule
\textbf{Approach} & \textbf{Signal} \\
\midrule
FRIDA (ours) & MIA/PIA on client updates \\
STD-DAGMM~\cite{lin2019free, zhu2021advanced} & DAGMM over updates \\
Feature-based~\cite{zhu2021advanced} & $\ell_2$ norm / standard deviation of updates \\
FRAD~\cite{wang2023frad} & DAGMM + contribution evaluation \\
Log-based~\cite{nguyen2024stake} & Client activity / stake logs \\
Incentive-based~\cite{lyu2020collaborative,xu2020reputation,wang2023pass} & Reputation / reward signals \\
\bottomrule
\end{tabular}
\end{table}

\subsubsection{Privacy Considerations}

Privacy-Preserving Federated Learning (PPFL) introduces challenges in detecting malicious clients and scoring honest ones due to the obfuscation of gradient updates. Most existing schemes rely on advanced cryptographic primitives such as Homomorphic Encryption (HE) for gradient comparison~\cite{so2020byzantine,liu2021privacy,ma2022shieldfl,miao2022privacy}. While these approaches achieve accuracy comparable to their plain-text counterparts, their significant computational overhead makes them impractical for real-world applications. Another line of PPFL employs more lightweight cryptographic solutions, such as SA~\cite{pejo2023quality,xhemrishi2023fedgt}, relying on group testing instead of fine-grained individual-level information. Although these approaches provide privacy by design, their accuracy typically falls short compared to other techniques. 

Client-level DP~\cite{pejo2022guide} presents another strategy to mitigate FR behavior in non-sybil settings by limiting a client's impact through gradient clipping. However, this defense introduces noise into the training process, delaying model convergence. As such, while DP may be effective against malicious attackers, its application to mitigate FR could inadvertently mimic the negative effects of FR itself.

\subsection{Privacy Attacks}

Privacy concerns in FL primarily stem from the potential exposure of participants' local data, which may include personal, sensitive, or proprietary information. While FL inherently offers some privacy protection by sharing only the model updates, numerous studies have shown unintended information leakage through various attack strategies~\cite{mothukuri2021survey}. Although privacy-preserving techniques like SA and DP exist, they often come with trade-offs, such as reducing model accuracy or preventing misbehaviour detection.

\subsubsection{Membership Inference Attack}

ML models tend to exhibit different behaviours for members and non-members of the training dataset, often not exclusively due to overfitting~\cite{yeom2018privacy}. Since training data may include sensitive personal information (e.g., patients' medical conditions), learning such information via membership inference attacks ~\cite{hu2022membership} can be considered a privacy violation. 

We can classify MIAs as black-box or white-box, depending on the access to the model parameters. It is well known that white-box provides no benefit compared to black-box~\cite{sablayrolles2019white}. However, this has been questioned recently~\cite{li2024unveiling}. 

The first MIA used shadow models~\cite{shokri2017membership} that mimic the behavior of the target model. Since the membership information for these shadow models is known, an attack model can be trained on them to determine if a sample was used in the training of the target model. Later, other black-box attacks were introduced that do not require training additional models. One such attack relies on prediction loss~\cite{yeom2018privacy} and uses a threshold to detect training samples.

Recently, the authors of~\cite{li2023effective} observed that gradients of different instances are orthogonal in overparameterized models. They argued that this is due to the sparsity of the high-dimensional gradients, which behave similarly to independent random isotropic vectors. Based on this observation, they develop an attack based on cosine similarity, which can reveal membership if a target sample's gradients are not orthogonal to the client's update, which is the average of the training gradient samples. In this paper, we utilize the loss-based attack by Yeom et al. ~\cite{yeom2018privacy} and the cosine-based attack by Li et al. ~\cite{li2023effective}, referring to them as $\mathcal{M}^Y$ and $\mathcal{M}^L$. Both attacks are detailed in \ref{sec:appendix_mia}.

\subsubsection{Property Inference Attack}

PIA, instead of focusing on individual-level information, aims to infer sensitive properties of the entire local training dataset~\cite{ateniese2015hacking,wang1910eavesdrop}. Typically, it utilizes meta-classifiers trained on shadow models~\cite{ganju2018property,parisot2021property}; it simulates training on shadow datasets with and without a property and then trains a meta-classifier to differentiate between the two.

Many attacks in the federated setting rely on gradient updates, such as those in \cite{zhou2021property,wainakh2021user,dai2024decaf}, where the inferred property is the client's label distributions. In this paper, we utilize the attacks by Wainakh et al.~\cite{wainakh2021user} and Dai et al.~\cite{dai2024decaf}, referring to them as $\mathcal{P}^W$ and $\mathcal{P}^D$, respectively. These attacks rely on the client's gradients corresponding to the last layer, the model from the previous round, and the number of samples used by the client to infer the label distribution of the clients. Formally, using the notations from Table~\ref{tab:notations}, $L_n^t=\mathcal{P}(G_n^t,M^{t-1},|D_n|)$. We detail both attacks in \ref{sec:appendix_pia}.

\section{Model}
\label{sec:mod}

\begin{figure}[!b]
    \centering
    \includegraphics[width=0.5\textwidth]{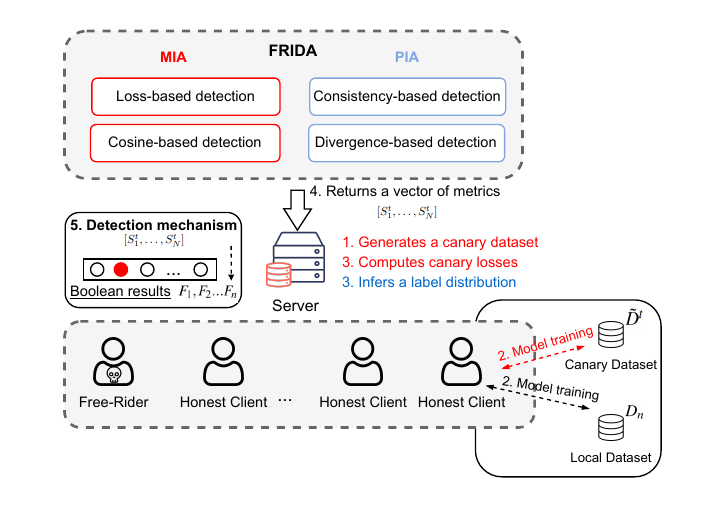}
    \caption{The FL process includes FR clients and the proposed detection metrics. Actions performed by the server for PIA mechanisms are shown in blue, while those for MIA mechanisms are shown in red.} 
    \label{fig:overview}
\end{figure}

This section details how FRIDA leverages MIA and PIA to detect free-riders. Figure ~\ref{fig:overview} provides an overview of the entire scenario. In each round, the server broadcasts the aggregated model, which honest clients further train on their local data, while FR clients apply one of the FR strategies. \\
\textit{In MIA-based detection}, the server shares a small dataset (100 samples as default), referred to as the canary set, with the clients in each round. In addition to local training, clients are required to train their local model on this canary set.\\ 
\textit{In both MIA and PIA detection}, beyond aggregating client updates, the server applies one of the four proposed FR detection mechanisms, each involving different steps. For instance, MIA-based detection requires the server to compute the corresponding canary losses or cosine similarity metrics. In contrast, PIA requires training on auxiliary data, which can be generated from noise, to infer the underlying client label distributions.

\begin{table}[!b]
\footnotesize	
    \centering
    \begin{tabular}{p{1.1cm}|p{6.5cm}}
        Symbol & Description \\
        \hline
        $N$ \& $T$ & Number of clients \& rounds\\
        $M^t_n \& G_n^t$ & Client $n$'s model \& gradient in round $t$ \\
        $M^t \& G^t$ & Aggregated model \& gradient in round $t$ \\
        $D_n$ \& $\tilde{D}$ & Dataset for client $n$ \& for the server \\
        $\mathcal{M}$ \&         $\mathcal{P}$ & Membership \& Property Inference Attack\\
        $\tau_M$\&$\tau_P$ & Threshold for MIA \& PIA to flag FR clients \\
        $c$ \& $l$ & Number of canaries \& labels \\
        $L_n^t$ & Inferred label dist. for client $n$ in round $t$ \\
        $\mathcal{F}$ & Applied FR technique (instead of training) \\
    \end{tabular}
    \caption{Notations used in the paper.}
    \label{tab:notations}
\end{table}

\subsection{Notations}

FL enables $N$ clients to collaboratively train a shared model $M(\theta)^t$ with parameters $\theta=\{\theta_1,\dots,\theta_m\}$ over $T$ rounds ($1 \leq t \leq T$). During local training, each client $n$ minimizes its empirical loss $\min_\theta \sum_{s\in D_n}\mathcal{L}(s,\theta)$, using $s$ samples from their local dataset $D_n$. The resulting local models $M_n^t$ are aggregated without requiring to share private datasets. For clarity, we omit the explicit mention of the parameters $\theta$ in subsequent sections. Table~\ref{tab:notations} summarizes the notations used in the paper.

\subsection{Assumptions and privacy discussion}

\begin{table}[!b]
\footnotesize	
    \centering
    \begin{tabular}{c|c|c|c}
        \multirow{2}*{Attack}&\multirow{2}*{Method} & \multicolumn{2}{c}{Training Requirement} \\
        \cline{3-4}
         && Server & Client \\
        \hline
        \multirow{2}*{MIA} & Loss &  Nothing & Canaries \\
        & Cosine & Canaries & Canaries \\
        \hline
        \multirow{2}*{PIA} & Consistency & Noise & Nothing \\
        & Diversity & Noise & Nothing \\
    \end{tabular}
    \caption{Training requirements for the four proposed FR detection mechanisms. }
    \label{tab:compare}
\end{table}

Our detection methods assume compliance from either client or server side. For MIA-based detection, we assume that the server has access to a small dataset, and that clients are compliant to train over this dataset. On the other hand, PIA-based detection does not need any subset of real data nor extra client training, but assumes that the server has white-box access to the global model. Table~\ref{tab:compare} summarizes the assumptions and requirements associated with each proposed FR detection mechanism.

FRIDA requires the ability to inspect individual client updates, a foundational requirement it shares with many BFT and CE schemes. This positions our framework in contrast to methods that prioritize client anonymity through privacy-preserving techniques, a trade-off we address in Section \ref{sec:discussion}. However, our strategy is highly practical in specific contexts such as cross-silo decentralized environments, where ensuring fair contribution is a critical requirement.

Nonetheless, we propose privacy-preserving configurations that enable the deployment of FRIDA without compromising clients' privacy. One such design is a two-server architecture which provides MIX-like~\cite{chaum1981untraceable} functionality, breaking the link between clients' IDs and their updates, as demonstrated in ShieldFL~\cite{ma2022shieldfl}. In this setup, client identifiers and model updates are processed by separate servers, allowing the server to use the gradients to detect FR behavior while preserving client privacy. This adds a single inter-server transfer per round carrying all updates $O(\theta N)$, and linear-time shuffle at the servers. There is no extra client-side work and no change to the per-round detection costs reported in Table \ref{tab:overhead}.

Alternatively, the clients can utilize local DP~\cite{pejo2022guide} which grants their data samples a privacy shield from the server. While this degrades model utility, it also mitigates the effectiveness of privacy attacks \textemdash such as MIA and PIA \textemdash on which FRIDA relies. We investigate the impact of local DP in our ablation study in Section \ref{sec:ablation}, and demonstrate that PIA-based detection remains effective in this scenario. 

\subsection{Free-riding Strategies}

% Many related research papers employ simple FR strategies, such as using zero or the previous global gradient for model updates. These basic FR attacks are easily detected, 

In our study, we consider the FR strategies by Fraboni et al.~\cite{fraboni2021free}, Lin et al.~\cite{lin2019free} and Zhu et al.~\cite{zhu2021advanced}, referred to as $\mathcal{F}^F$, $\mathcal{F}^L$ and $\mathcal{F}^Z$, respectively. These attacks simulate SGD behavior by adding Gaussian noise to specific components of the global model $M^{t}_n\sim \mathcal{N}(\mu,\sigma)$, such that $\mu_F=M^{t-1}$ for $\mathcal{F}^F$, $\mu_L=M^{t-1}+G^{t-1}$ for $\mathcal{F}^L$, and $\mu_Z=M^{t-1}+G^{t-1}$ scaled by a non-trivial factor for $\mathcal{F}^Z$. The noise variance $\sigma$ is adjusted based on the empirical variance of the global gradient $G^{t-1}$, along with additional multiplicative factors and time-dependent variables.

\subsubsection{Threat Model}

In this work, we consider anonymous FR attackers with no access to data, which is common in sensitive FL scenarios. By default, we consider free-riders that are active at every round; because detection is performed independently per round, our evaluation naturally covers adaptive strategies that appear only in specific rounds. 

Regarding MIA-based detection, FRIDA requires client compliance to train on a small canary set in addition to their own local data. Thus, we assume that FR clients lack the computational resources to train on the canary dataset, as it would reduce the practicality of such a lightweight attack. Therefore, their capability is to craft per-round updates from public signals (e.g., prior global model or aggregated gradient) without access to local data or canary training. Nevertheless, we study how MIA-based methods perform when free-riders are adaptive in Section \ref{sec:discussion}. 

\subsubsection{FR detection} 

At each round $t$, we use the scores $[S_1^t,\dots,S_N^t]$ from our detection methods to flag potential FR clients. Specifically, we apply a \textit{$z$-score} test, which identifies outliers by measuring the difference from the mean, normalized by the standard deviation. First, we calculate the $z$-score for every client, as shown in Equation~\ref{eq:outlier}, and then flag a client as FR if their score exceeds a threshold $\tau$. 
\begin{equation}
    \label{eq:outlier}
    Z_n^t=\frac{\left|S_n^t-\text{avg}_n\left(S_n^t\right)\right|}{\text{std}_n\left(S_n^t\right)}
\end{equation}

% LIMITATIONS, not here
% The main drawback of these anomaly detection schemes is their fidelity: they are compromised when there are no anomalies. Although cosine-based detection bypasses this limitation by considering each client individually, we also explore a slight variation of our loss-based detection methods in the following section. For our experiments, we set the threshold to the standard value to ensure a fair comparison across detection techniques; it is set independently of the proposed solutions and used consistently across all $z$-tests. 

% \begin{algorithm}[!b]
%     \caption{$z$-test for FR detection.}
%     \label{alg:threshold}
%     \textbf{Input: }
%     $[S_1^t,\dots,S_N^t]$ (scores from MIA or PIA), $\tau$ \\
%     \textbf{Output: }
%     binary vector flagging FR clients $[F_n^t]_1^N$
%     \begin{algorithmic}
%         \State $F_n^t \gets [0,\dots,0]$
%         \For{$n \in [1,\dots,N]$}
%             \State $Z_n^t \gets \frac{\left|S_n^t-\text{avg}_n\left(S_n^t\right)\right|}{\text{std}_n\left(S_n^t\right)}$
%             \If {$Z_n^t > \tau$}
%                 \State $F_n^t \gets 1$
%             \EndIf
%         \EndFor
%     \end{algorithmic}
% \end{algorithm}

\subsection{FR detection via MIA}

The membership inference attack (MIA) is a privacy attack used to determine whether a particular sample was included in the training data. Following prior works, we assume that the server has access to a small canary subset of labeled data $\tilde{D}$. In this regard, we re-purpose loss-based and a cosine-based MIA strategies to detect FR clients using the canary dataset. Both approaches follow a general two-step procedure: a MIA $\mathcal{M}$ is utilized to obtain detection scores, which are then compared with a threshold $\tau_M$ to distinguish between FR and honest clients.

Our strategy for identifying FR clients with MIA involves providing $c$ canary samples to each client. The clients train their local models over the canary samples after their local training. Formally, in round $t$, an honest client $n$ first trains the model $M^{t-1}$ on their dataset $D_n$ by minimizing the loss $\min_\theta\sum_{s\in D_n}\mathcal{L}(s,M^{t-1}(\theta))$. Then, the client trains the model with a subset $\tilde{D}^t\subset\tilde{D}$ of size $c$ from the canary set, minimizing the loss $\min_\theta\sum_{s\in \tilde{D}^t}\mathcal{L}(s,M^{t-1}(\theta))$. While the canary samples are fixed for all clients, a different subset of canary samples is chosen for each round.

Then, the server employs $\mathcal{M}$ to detect the clients who did not use the canary samples for training, resulting in a set of scores $S_n^t(s)$. Using both the original global model $M^{t-1}$ and the locally computed gradient $G_n^t$, the server uses $\mathcal{M}$ to collect a matrix of scores $S_{N\times c}$, which is then analyzed to detect FR clients.

\subsubsection{Loss-based MIA}

Here, we adapt $\mathcal{M}^Y$~\cite{yeom2018privacy} to our canary setup, formally defined in Equation~\ref{eq:MIA_loss}. Upon receiving the local models $M_n^t$ from clients, the server computes the canary losses on these updates, resulting in the matrix $S_{N\times c}^t$. Detection is performed by flagging a client as FR if its canary loss is an outlier according to the $z$-score mechanism. The full pseudo-code of the FR detection is presented in Algorithm~\ref{alg:loss}, where $[\cdot]_1^N$ means a vector enumerating $n$ in $[1,N]$. 

\begin{align}
    \label{eq:MIA_loss}
    S_n^t = \mathcal{M}^Y(M^{t-1},G_n^t,s) = \mathcal{L}(s,M_n^t(\theta)) 
    % \label{eq:MIA_threshold}
    % \sum_{i=1}^c S_{n;i}^t \geq \tau_M = \sum_{i=1}^c\tilde{S}_i^t
\end{align}

\begin{algorithm}[!b]
    \caption{Loss-based FR detection.}
    \label{alg:loss}
    \textbf{Input: }
    $M^{t-1}, [G_n^t]_1^N$, $\tilde{D}^t$ for MIA $\mathcal{M}^Y$, $\tau_M$ for $z$-test $\mathcal{Z}$ \\
    \textbf{Output: }
    binary vector flagging FR clients $[F_n^t]_1^N$
    \begin{algorithmic}
        \State $\tilde{G}^t \gets \min_\theta\sum_{s\in \tilde{D}^t}\mathcal{L}(s,M^{t-1}(\theta))$
        % server-side only
        %\State ${\tau}_M \gets \sum_{s\in \tilde{D}^t}\mathcal{M}^Y(M^{t-1},\tilde{G}^t,s)$
        \For{$n \in [1,\dots,N]$}
            \State $F_n^t \gets 0$
            \For{$s \in \tilde{D}^t$}
                \State $S^t_{n;s} \gets \mathcal{M}^Y(M^{t-1},G_n^t,s)$
            \EndFor            
            \State $F_n^t \gets \mathcal{Z}([\sum_{s\in \tilde{D}^t}S_{n;s}^t]_1^N,\tau_M)$
        \EndFor
    \end{algorithmic}
\end{algorithm}

\subsubsection{Cosine-based MIA}

% We adapt $\mathcal{M}^L$~\cite{li2023effective} to adapted to our canary setup as detailed in Equation~\ref{eq:MIA_cos}. Note, we slightly overload the term $\mathcal{M}^L$ as in Algorithm~\ref{alg:cosine} we replace the input $s$ with $\tilde{G}_s^t$. 

Also, we adapt $\mathcal{M}^L$~\cite{li2023effective} to FR detection, formalized in Equation~\ref{eq:MIA_cos}. During round $t$, the server computes gradients for both the canary set samples $\tilde{D}^t$ and a random complement set of equal size $\tilde{D}^t_C\in\tilde{D}\backslash\tilde{D}^t$. These gradients are denoted as $[\tilde{G}_{s_1}^t,\dots,\tilde{G}_{s_c}^t]$ and $[\tilde{G}_{s^\prime_1}^t,\dots,\tilde{G}_{s^\prime_c}^t]$, respectively, where $s\in\tilde{D}^t$ and $s^\prime\in\tilde{D}^t_C$. The gradients are compared using cosine similarity with those from the clients. This yields score vectors $S^t_n$ and $\tilde{S}^t_n$ for each client $n$. Since gradients of different samples are orthogonal in overparameterized models, and the cosine similarities of the gradients are normally distributed~\cite{li2023effective}, we can employ a $t$-test with parameter $\tau_M$ to compare the means of the distributions behind $S_N^t$ and $\tilde{S}_n^t$~. If the two distributions are not statistically different, the client is flagged as FR, indicating that the client has not trained over the canary set. The pseudo-code for this mechanism is in Algorithm~\ref{alg:cosine}.
\begin{align}
    \label{eq:MIA_cos}
    S_n^t = \mathcal{M}^L(M^{t-1},G_n^t,s) = CosSim(\tilde{G}_s^t;G_n^t)
\end{align}

\begin{algorithm}[!b]
    \caption{Cosine-based FR detection.}
    \label{alg:cosine}
    \textbf{Input: }
    $[G_n^t]_1^N, M^{t-1}, \tilde{D}^t$ for MIA $\mathcal{M}^L$, $\tau_M$ for $t$-test $\mathcal{T}$ \\
    \textbf{Output: }
    binary vector flagging FR clients $[F_n^t]_1^N$
    \begin{algorithmic}
        \For{$n \in [1,\dots,N]$}
            \For{$s \in \tilde{D}^t$}
                \State $\tilde{G}^t_s \gets \min_\theta\mathcal{L}(s,M^{t-1}(\theta))$
                \State $S^t_{n;s} \gets \mathcal{M}^L(M^{t-1}, G_n^t, \tilde{G}^t_s)$
            \EndFor
            \For{$s^\prime \in \tilde{D}^t_C$}
                \State $\tilde{S}^t_{n;s^\prime} \gets \mathcal{M}^L(M^{t-1}, G_n^t, \tilde{G}^t_{s^\prime})$
            \EndFor
        \EndFor
        \For{$n \in [1,\dots,N]$}
            \State $T_n^t \gets \mathcal{T}(S^t_n, \tilde{S}^t_{n})$
            \State $F_n^t \gets 0$
            \If{$T_n^t \geq \tau_M$}
                    \State $F_n^t \gets 1$
            \EndIf
        \EndFor
    \end{algorithmic}
\end{algorithm}

\subsection{FR detection via PIA}

The property inference attack (PIA) is a privacy attack designed to infer sensitive properties of the local datasets. In this paper, we use $\mathcal{P}^W$~\cite{wainakh2021user} and $\mathcal{P}^D$~\cite{dai2024decaf} to estimate the label distribution of the client's local datasets. After round $t$, the PIA $\mathcal{P}$ receives the aggregated model $M^{t-1}$ from the previous round, along with clients' gradients $G^t_n$ and dataset sizes $|D_n|$. It then estimates each client label distribution $L^t_n$, as described in Equation~\ref{eq:PIA}. Based on the inferred label distribution $L^t_n$, a score $S^t_n$ is computed for each client, used to identify FR clients using the z-score test and a threshold $\tau_P$. To generate these scores, we propose two techniques: \textit{consistency-based} and \textit{diversity-based}, presented in Algorithm~\ref{alg:consistency} and Algorithm~\ref{alg:diversity}, respectively.
\begin{equation}
    \label{eq:PIA}
    L_n^t=\mathcal{P}(G_n^t,M^{t-1},|D_n|)
\end{equation}

\subsubsection{Consistency-based detection}  

Intuitively, the label distributions for the same clients across different rounds $t$ should remain consistent, such that $\forall \hspace{0.1cm}t\in[2,\dots,T], L_n^t \sim L^{t-1}_n $. While the inherent noise of the training process introduces variations on the label distribution across rounds, the magnitude of these variations is similar for all honest clients. Building on this observation, our consistency-based FR detection relies on the following premise: depending on the FR strategy and noise added, the differences between the label distributions for FR clients across rounds are either larger or smaller than the differences for honest clients. In this regard, we calculate the average Euclidean distances between current and previous inferred distributions, as shown in Equation $\ref{eq:PIA_consistency_score}$.

\begin{align}    
    \label{eq:PIA_consistency_score}
    S_n^t=Euclid(L_n^t,L_n^{t-1})
\end{align}

\begin{algorithm}[tb]
    \caption{Consistency-based FR detection.}
    \label{alg:consistency}
    \textbf{Input: }
    $[G_n^t]_1^N, M^{t-1},[|D_n|]_1^N$ for $\mathcal{P}$, $\tau_P$ for $\mathcal{Z}$, $[L_n^{t-1}]_1^N$ \\
    \textbf{Output: }
    binary vector flagging FR clients $[F_n^t]_1^N$
    \begin{algorithmic}
        \For{$n \in [1,\dots,N]$}
            \State $L_n^t \gets \mathcal{P}(G_n^t,M^{t-1},|D_n|)$
            \State $S_n^t \gets Euclid(L_n^t,L_n^{t-1})$
        \EndFor
        \State $F_n^t \gets \mathcal{Z}([S_n^t]_1^N,\tau_P)$
    \end{algorithmic}
\end{algorithm}

\subsubsection{Diversity-based detection} 

Also, we introduce a diversity-based FR detection, which is based on the following two hypotheses. First, if a FR client update is mostly driven by noise, the inferred label distribution $L_n^t$ will be nearly uniform, such that $L_n^t \sim \overline{L} = [l^{-1},l^{-1},\dots,l^{-1}]$, where $l$ is the number of labels. On the other hand, if FR clients updates mostly rely on the previous global update, the inferred distribution will be close to the global label distribution $L^t$, defined as 
\begin{equation}
    L^t = \sum_{i=1}^N\frac{\mathcal{P}(G_i^t,M^{t-1},|D_i|)}{N}.
\end{equation}

We approximate the inferred label distribution for all clients as a linear combination of $\overline{L}$ and $L^t$, with $\alpha$ and $\beta$ optimized to minimize the \textit{mean square error} (MSE) $\min_{\alpha,\beta} [L^t_n - (\alpha \cdot L^t + \beta \cdot \overline{L})]^2$. Based on the obtained optimal values $\alpha^*$ and $\beta^*$, the diversity score is defined as the logarithm of the MSE (see Equation~\ref{eq:PIA_dist_diversity}), which is a common approach in outlier detection. The complete diversity-based detection mechanism is shown in Algorithm~\ref{alg:diversity}.
\begin{align}
    \label{eq:PIA_dist_diversity}
    S_n^t=\log[L^t_n - (\alpha^* \cdot L^t + \beta^* \cdot \overline{L})]^2
\end{align}

\begin{algorithm}[tb]
    \caption{Diversity-based FR detection.}
    \label{alg:diversity}
    \textbf{Input: }    
    $[G_n^t]_1^N, M^{t-1},[|D_n|]_1^N$ for PIA $\mathcal{P}$, $\tau_P$ for $z$-test $\mathcal{Z}$ \\
    \textbf{Output: }
    binary vector flagging FR clients $[F_n^t]_1^N$
    \begin{algorithmic}
        \For{$n \in [1,\dots,N]$}
            \State $L_n^t \gets \mathcal{P}(G_n^t,M^{t-1},|D_n|)$
        \EndFor
        \State $L^t \gets \frac{\sum_{i=1}^N{L_n^t}}{N}$ ; $\overline{L} \gets [l^{-1}]_1^N$
        \For{$n \in [1,\dots,N]$}
            \State $\alpha^*, \beta^* \gets \min_{\alpha,\beta} [L^t_n - (\alpha \cdot L^t + \beta \cdot \overline{L})]^2$
        \State $S_n^t \gets \log[L^t_n - (\alpha^* \cdot L^t + 
        \beta^* \cdot \overline{L})]^2$
        \EndFor
        \State $F_n^t \gets \mathcal{Z}([S_n^t]_1^N,\overline{\tau}_P,\underline{\tau}_P)$
    \end{algorithmic}
\end{algorithm}

\subsection{Computational Overhead}
\label{sec:comp_overhead}

We summarize the computational cost and runtime of the proposed algorithms and feature-based methods in Table~\ref{tab:overhead} and Table \ref{tab:runtime} considering compute and memory costs. Here, $h_i$ denotes the size (height) of layer $i$ in the neural network, $d$ represents the depth (the number of layers), and $\Theta=\sum_{i=2}^{d}{h_i \cdot h_{i-1}}$ indicates the total number of parameters. 

\subsubsection{MIA-based detection}

Loss-based detection requires a forward pass for each client with the canary dataset to obtain the canary losses. This operation takes $O(N \cdot |\tilde{D}| \cdot \Theta )$ time, while the $z$-score calculation is performed in $O(N)$ time. 

On the other hand, cosine-based detection requires a backwards pass to obtain the member and non-member gradients, which takes $O(|\tilde{D}| \cdot \Theta)$ time. The computation of cosine similarity for each batch and client adds an additional $O(N \cdot |\tilde{D}| \cdot \Theta)$ time. Finally the $t$-test involves comparing distributions for each client, where the number of samples equals the number of batches, resulting in $O(N \cdot |\tilde{D}|)$ time.

\subsubsection{PIA-based detection}
When using PIA-based solutions, we must consider three main components: the PIA itself (training and inference), the calculation of metrics (consistency or diversity), and outlier detection. Regarding training, both $\mathcal{P}^W$ and $\mathcal{P}^D$ requires $O(|\tilde{D}| \cdot \Theta)$ time, while inferring the properties using the weights of the last layer for each client takes $O(h_d \cdot h_{d-1} \cdot N)$. Note that the last layer has one neuron per label, so $l=h_d$. Finally, calculating both metrics involves operations that are linear in the number of labels per client, similarly to the outlier detection mechanism. 

\subsubsection{Comparison with feature-based detection}

Feature-based baselines are computationally lightweight, as computing each statistic requires a single pass over a flattened client update, so the per-round cost is $O(N \cdot \Theta)$ with memory cost $O(\Theta)$. In contrast, DAGMM-based detectors \cite{lin2019free, wang2023frad} learn to reconstruct each client’s flattened model update with an autoencoder, followed by fitting a Gaussian Mixture Model to compute anomaly scores. The computational overhead of DAGMM can be decomposed on training the autoencoder across $N$ client updates of size $\Theta$, which can be approximated as $O(N \cdot \Theta)$ per round, and the GMM fitting adds $O(k \cdot N \cdot d^2)$, where $d$ is the latent dimension and $k$ the number of components. Moreover, the memory footprint within the server grows with the number of client updates $O(N \cdot \Theta)$, which scales poorly for medium- to large-sized models and larger numbers of clients. 

\begin{table}[!b]
    \small
    \centering
    \renewcommand{\arraystretch}{1.4}
    \setlength{\tabcolsep}{4pt}
    \begin{tabular}{lcc}
        \toprule
        \textbf{Method} & \textbf{Computation-cost} & \textbf{Memory-cost} \\
        \hline
        Loss (MIA)        & $O(N\,|\tilde{D}|\,\Theta)$                          & $O(\Theta + |\tilde{D}| + N)$ \\
        Cosine (MIA)      & $O(N\,|\tilde{D}|\,\Theta)$                          & $O(\Theta + |\tilde{D}| + N)$ \\
        PIA & $O(|\tilde{D}|\,\Theta + N\,h_d h_{d-1})$            & $O(\Theta + |\tilde{D}| + N\,h_d h_{d-1})$ \\
        \midrule
        Feature (L2/STD/Cos) & $O(N \Theta)$ & $O(\Theta + N)$ \\
        STD-DAGMM & $O(N\,d\;\Theta\ + k\,N\,d^{2})$ & $O(\Theta\,d + \Theta\,N + k\,d^{2})$\\
        \bottomrule
    \end{tabular}
    \caption{Per-round compute and peak working-set memory.}
    \label{tab:overhead}
\end{table}

% \begin{table}[h] 
% \small 
% \centering 
% \renewcommand{\arraystretch}{1.3} 
% \setlength{\tabcolsep}{4pt} 
% \begin{tabular}{lcc} 
% \toprule \textbf{Method} & \textbf{Time / round (s)} \\ \midrule 
% Loss–MIA & 2.58 
% \\ Cosine–MIA & 1.97 
% \\ PIA & $0.30$ \\ \midrule Feature-based & $0.19$ 
% \\ STD-DAGMM & $16.62$ 
% \\ \bottomrule \end{tabular} \caption{Per-round server detection time and peak GPU memory under 100 clients and 20 free-riders across detection methods. } \label{tab:runtime} \end{table}

\begin{table}[h]
\footnotesize
\centering
\renewcommand{\arraystretch}{1.3}
\setlength{\tabcolsep}{4pt}
\begin{tabular}{lccccc}
\toprule
& \textbf{Loss} & \textbf{Cosine} & \textbf{PIA} & \textbf{Feature-based} & \textbf{STD-DAGMM} \\
\midrule
\textbf{Time / round (s)} & 2.58 & 1.97 & 0.30 & 0.19 & 16.62 \\
\bottomrule
\end{tabular}
\caption{Per-round server detection time under 100 clients and 20 free-riders across detection methods. MIA methods use a canary size $|\tilde{D}| = 100$}
\label{tab:runtime}
\end{table}

\section{Experiments}
\label{sec:exp}

In this section, we evaluate the performance of FRIDA across a range of experimental settings and compare it against feature-based FR detection methods, which serve as the foundation of most defenses by indirectly capturing FR behavior.

% Each Figure represents a specific combination of model, dataset, number of clients, data distribution and FR strategy. By default, we examine AlexNet trained on CIFAR-100, distributed across eight clients (conducting three epochs of canary training in the case of MIA), where one client is a FR client.  

\subsection{Experimental set-up}

\paragraph{\textbf{Datasets and Models}} We use \textit{CIFAR-10} and \textit{CIFAR-100}~\cite{krizhevsky2009learning} as image classification baseline datasets. These datasets comprise 60000 32x32 color images distributed across 10 and 100 classes, respectively. For the model architecture, we utilize \textit{AlexNet} \cite{krizhevsky2012imagenet}, a 62M parameters deep convolutional network, and VGG-19, included in Table \ref{tab:ablation}. All experiments are repeated five times; we report the empirical mean and standard deviation of the results. 

\paragraph{\textbf{FL Parameters}} Federated training follows FedAvg~\cite{mcmahan2017communication} as the aggregation rule. We consider both a cross-silo setting with 8 clients, and a cross-edge scenario with 100 clients. We evaluate three FR strategies: $\mathcal{F}^F$, $\mathcal{F}^L$ and $\mathcal{F}^Z$. The FL setup is simulated by assigning 4000 samples to each client, drawn from IID and non-IID distributions. For the non-IID scenario, we use a Dirichlet distribution with parameter $\alpha=1.0$ to simulate moderate heterogeneity and $\alpha=0.1$ for more significant differences. Also, a representative portion of IID data is allocated to the server for MIA-based detection. 

The FL environment is simulated with \textit{Flower}~\cite{beutel2020flower} and PyTorch. We focus on the worst-case scenario by limiting local epochs to one, as a larger number of epochs leads to more pronounced differences between honest and FR clients. We set the local batch size to 64 and use stochastic gradient descent (SGD) with momentum $0.9$, weight decay $10^{-5}$, and learning rate $0.01$. All experiments are executed on a NVIDIA H100 64GB GPU. 

\paragraph{\textbf{FR Parameters}} We configure the $\mathcal{F}^{F}$ free-rider strategy using a scaling factor $\alpha = 1$ and noise standard deviation $\sigma = 1$. For the $\mathcal{F}^{Z}$ FR strategy, we add noise to 60\% of the model parameters, and the constant $C$ is defined as the ratio between the $\ell_2$ norm of the model update $\| \mathcal{M}^\text{t} - \mathcal{M}^\text{0} \|_2$, where $\mathcal{M}^\text{0}$ represents the initial model, and the $\ell_2$ norm of the current model $\| \mathcal{M}^\text{t} \|_2$.

\paragraph{\textbf{MIA Parameters}} For canary-based MIA, at each round $t$, we randomly sample canary samples from $\tilde{D} \setminus \bigcup_{i=1}^{t-1} \tilde{D}^i$, setting $|\tilde{D}^t|$ to 100. These samples differ in each round but are consistent across all clients. After training on their datasets, we require the clients to train on the canary set for a specified number of epochs. In the cosine-based attack, we set the standard detection thresholds $\tau_M=0.05$ for the $t$-test, while for loss-based detection we use $\tau_M=1$ for the $z$-test. Finally, for visualization purposes, we smooth the cosine similarity distributions by averaging over a window of $10$ time-steps. 

\paragraph{\textbf{PIA Parameters}} For PIA, the auxiliary dataset consists of $20$ samples per label, with each sample generated from noise. The noise is sampled from a Gaussian distribution and transformed the same way as the training samples are prepared. The auxiliary gradient computation was conducted with a batch size of $32$. For the detection thresholds, we utilized the same $\tau_P=1$ threshold. 

\paragraph{\textbf{Feature-based baselines}} For feature-based outlier detection baselines, we compute per round: (i) the $\ell_2$ norm of each client’s update, (ii) the standard deviation of each client’s update, and (iii) the cosine similarity between each client’s update and a randomly selected client.

\paragraph{\textbf{Decision policy (thresholds)}} For all $z$-score decisions (feature baselines, STD-DAGMM, loss-MIA, PIA-consistency, PIA-diversity) we fix the threshold to $\tau=1$ standard deviation. For cosine-based MIA, we perform a per-client two-sample Student’s $t$-test with a two-sided significance level $\alpha=0.05$. This allows for a uniform decision policy across detectors and avoids threshold tuning per setup and dataset. We tested $\tau \in \{0.5,1,2\}$ and $\alpha \in \{0.01,0.05,0.1\}$, with $\tau=1$ and $\alpha=0.05$ yielding the best balance between detection and false positives.

\paragraph{\textbf{STD-DAGMM}} We implement STD-DAGMM \cite{lin2019free}, which augments the DAGMM feature vector with the standard deviation of client updates. This detector forms the outlier-detection backbone used in FRAD \cite{wang2023frad}. Our implementation uses a fully connected autoencoder using hidden-layer sizes (128, 64, 32), and a Gaussian mixture model with $K=3$ components. Following \cite{wang2023frad}, we set the loss weights to $\lambda_1 = 0.1$ and $\lambda_2 = 0.005$. 

\subsection{MIA Results}

In our default MIA experimental scenario, we assume that honest clients train over the canary set for three epochs.

\subsubsection{Loss-based detection}

First, we evaluate the performance of Algorithm~\ref{alg:loss} based on Yeom's loss-based MIA. In Figure~\ref{fig:loss_values}, we show that the loss values of the canaries computed on the honest clients (blue, orange, green) and FR clients (red) are different. This difference becomes more pronounced as training progresses, as the impact of training over the canary set remains stable while the gradients for their training set continue to decrease due to convergence. While one canary epoch is sufficient to ensure a separation between honest and FR clients, more canary epochs further increase this separation, improving the accuracy of the detection mechanism.

% This trend is also reflected in Figure~\ref{fig:MIA_loss_multiround}, where the F1-score improves across training rounds. 

\subsubsection{Cosine-based detection}

To verify the effectiveness of Algorithm~\ref{alg:cosine}, we show in Figure~\ref{fig:cos_values} the difference between the cosine-similarity distribution of the gradients given members (canary samples) and non-members for FR and honest clients. A client is flagged as FR if the means of the two distributions are statistically indistinguishable, determined by a Student's $t$-test, indicating that the client has not trained over the canary subset. For honest clients (right), the slight difference between the two distributions is sufficient to infer membership. 

% Similar to loss-based detection, as seen in Figure~\ref{fig:MIA_cos_multiround}, the accuracy of this method improves as training progresses.

\subsubsection{Canary effect}

\begin{figure}[tb]
    \centering
    \begin{subfigure}{0.3 \textwidth}
    \centering
    \includegraphics[width=\textwidth]{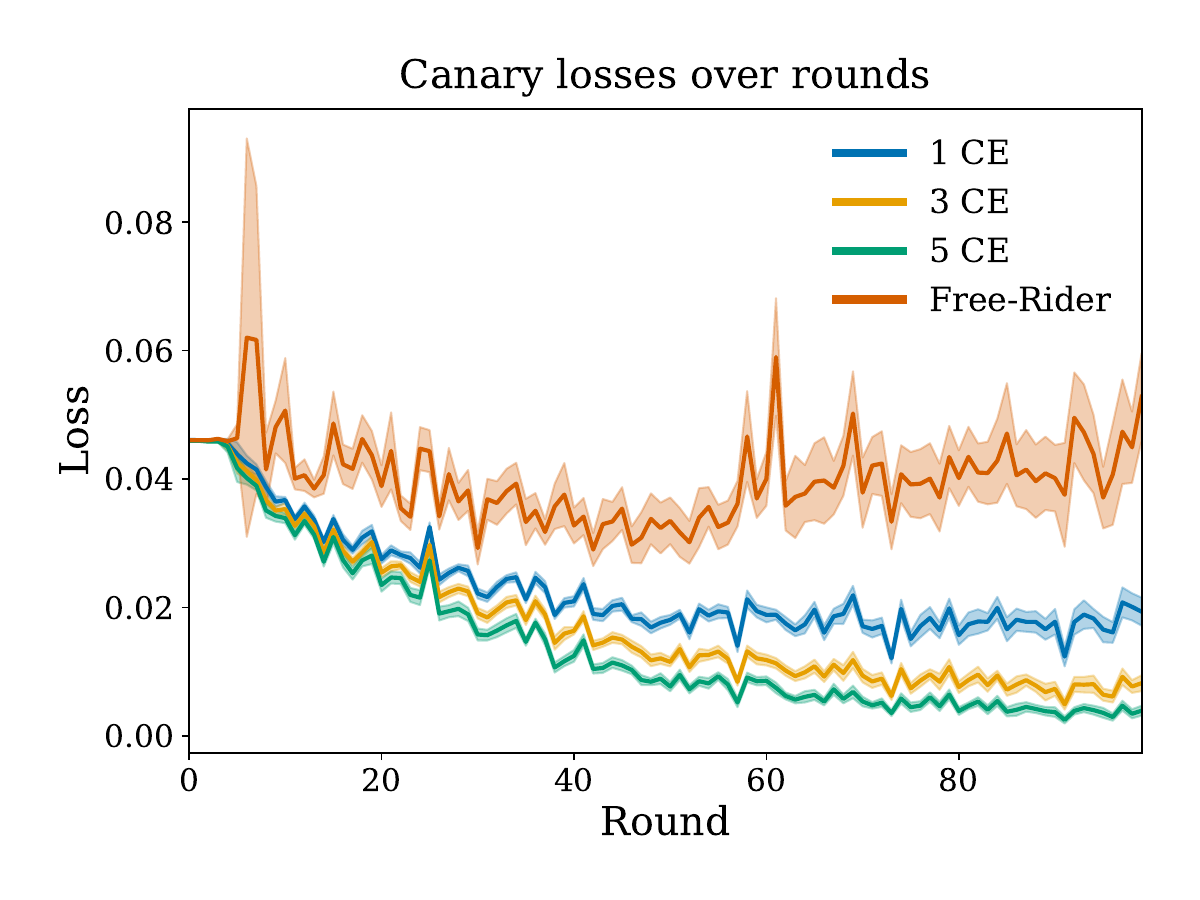}
    \caption{Round-wise loss values over canary samples for honest clients (blue, yellow, green) trained for 1/3/5 canary epochs (CE) and FR clients (red).}
    \label{fig:loss_values}
    \end{subfigure}%   <- note the percent sign!

    % \hspace{0.1\textwidth}%  <- explicit 10% gap
    \begin{subfigure}{0.45\textwidth}
    \centering
    \includegraphics[width=\textwidth]{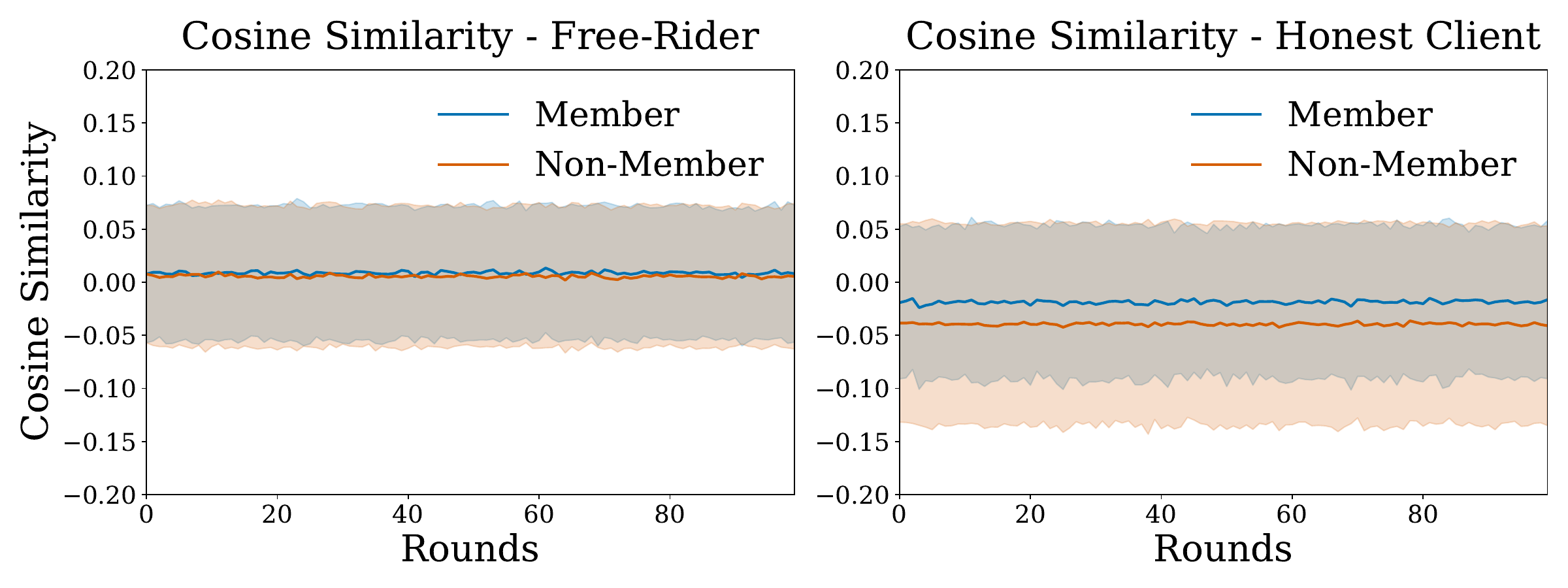}
    \caption{Round-wise cosine similarity values given canary (blue) and non-canary (red) samples for FR (left) and honest clients (right). For honest clients, the mean difference between the two distributions is statistically significant, indicating that they trained on the canary set.}
    \label{fig:cos_values}
    \end{subfigure}
    \caption{FR detection mechanisms using MIA with $\mathcal{F}^L$.}
    \label{fig:MIA}
\end{figure}

MIAs leverage the model's tendency to overfit to the training dataset to infer membership~\cite{yeom2018privacy}. This translates to our MIA-based detection mechanism, where greater overfitting to the canary dataset corresponds to improved FR detection performance. In Figure~\ref{fig:loss_values}, we observe that although FR clients are not training on the canary set, the canary loss is reduced in the early rounds. We argue that this results from improvements in the global model. However, after a certain number of rounds, the global model converges, and the canary loss is only reduced if the client has trained on the canary set.

We can induce targeted overfitting by increasing the number of canary training epochs. In Table~\ref{tab:canary}, we examine the effect of the number of canary epochs on detection performance, averaged over 100 rounds. This presents a balance between improving FR detection and maintaining overall training stability. Notably, FR clients significantly impact global training dynamics compared to canary-based detection mechanisms, as shown in Figure~\ref{fig:delay}. 

\begin{figure}[tb]
     \centering
     \includegraphics[width=0.45\textwidth]{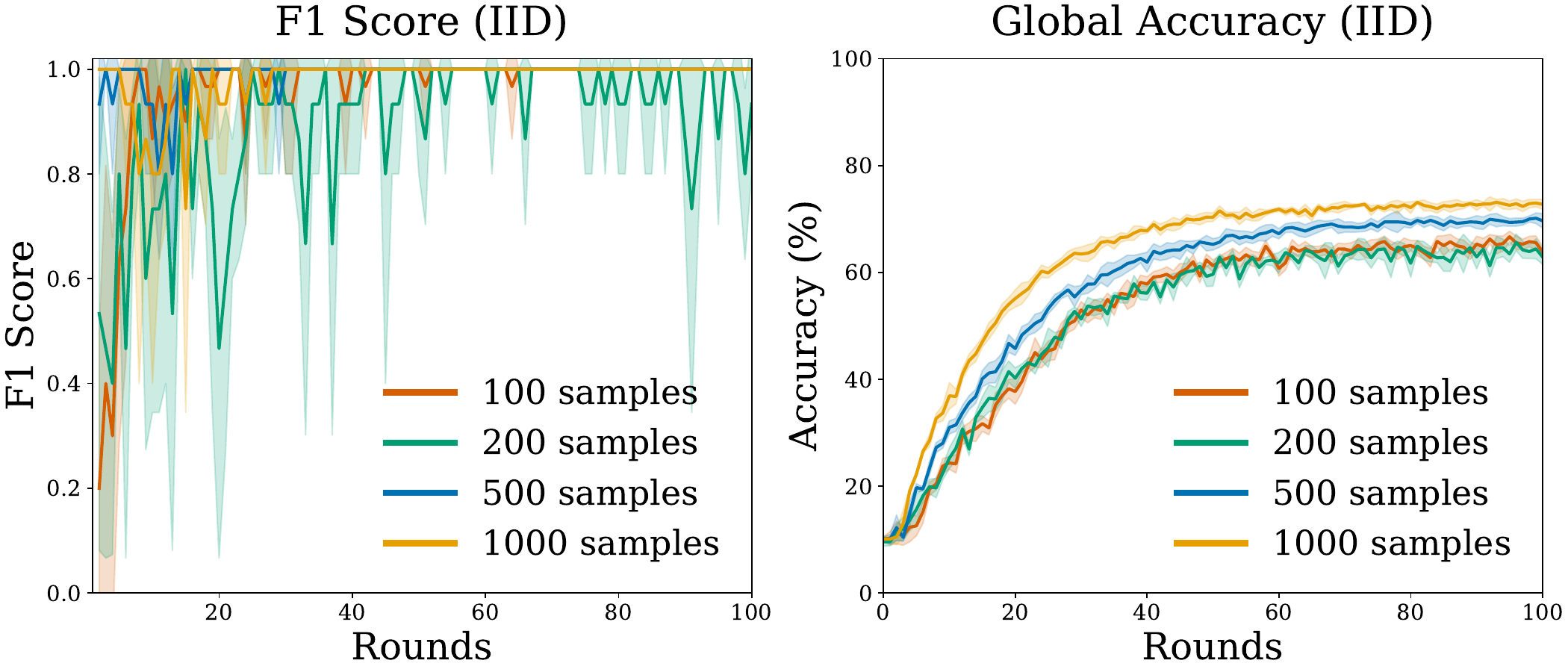}
     \caption{Round-wise comparison of the FR detection performance (left) and global accuracy (right) for varying numbers of canary samples. Larger canary sets implicitly increase the local training datasets, leading to faster global convergence.}
     \label{fig:more_canary_samples}
\end{figure}

We also study how increasing the canary samples improves detection performance. We can observe in Figure \ref{fig:more_canary_samples} that larger canary sizes lead to improved FR detection performance, as well as better global accuracy, since the client's training set is implicitly larger. In our experiments, we use 100 samples, as larger datasets may not always be available to the server.

% \subsubsection{Multi-round detection}

% The performance of the detection mechanism can be improved by aggregating knowledge from previous rounds. This improves detection of non-adaptive free-riders, which free-ride on every round, as considered in Section~\ref{sec:mod}. In this regard, we implement a majority-voting scheme where a client is flagged as FR if it has been identified as such more often than as honest within a given window of timesteps. Figures~\ref{fig:MIA_loss_multiround} and~\ref{fig:MIA_cos_multiround} show how this approach smooths and improves detection performance when combining knowledge from 20 consecutive rounds. This is particularly helpful in MIA-based mechanisms for improving performance in non-IID scenarios.

\begin{table}[tb]
\footnotesize	

    \centering
    \begin{subtable}{0.45\textwidth}
    \centering
    \begin{tabular}{|c|c|c|c|}
        \hline
        & \multicolumn{3}{|c|}{\textbf{Fraboni attack $(\mathcal{F}^F)$}} \\ \hline
        & \textbf{1 round} & \textbf{3 rounds} & \textbf{5 rounds} \\ \hline
        \textbf{IID}  & $0.60 \pm 0.03$ & $0.88 \pm 0.02$ & $0.99 \pm 0.01$ \\ \hline
        \textbf{N-IID 1.0}  & $0.65 \pm 0.04$ & $0.76 \pm 0.02$ & $0.90 \pm 0.00$ \\ \hline
        \textbf{N-IID 0.1}  & $0.36 \pm 0.05$ & $0.34 \pm 0.05$ & $0.51 \pm 0.02$\\ \hline
        & \multicolumn{3}{|c|}{\textbf{Lin attack $(\mathcal{F}^L)$}} \\ \hline
        & \textbf{1 round} & \textbf{3 rounds} & \textbf{5 rounds} \\ \hline
        \textbf{IID} & $0.80 \pm 0.03$ & $0.95 \pm 0.01$ & $0.98 \pm 0.01$ \\ \hline
        \textbf{N-IID 1.0}  & $0.60 \pm 0.02$ & $0.86 \pm 0.02$ & $0.92 \pm 0.01$ \\ \hline
        \textbf{N-IID 0.1} & $0.56 \pm 0.03$ & $0.48 \pm 0.03$ & $0.66 \pm 0.03$ \\ \hline
    \end{tabular}
    \caption{F1-scores of the loss-based detection. }
    \label{tab:loss_canary}
    \end{subtable}
    
    \vspace{0.1cm}
    \begin{subtable}{0.45\textwidth}
    \centering
    \begin{tabular}{|c|c|c|c|}
        \hline
        & \multicolumn{3}{|c|}{\textbf{Fraboni attack $(\mathcal{F}^F)$}} \\ \hline
        & \textbf{1 round} & \textbf{3 rounds} & \textbf{5 rounds} \\ \hline
        \textbf{IID}  & $0.83 \pm 0.06$ & $0.87 \pm 0.05$ & $0.94 \pm 0.01$ \\ \hline
        \textbf{N-IID 1.0} & $0.69 \pm 0.04$ & $0.78 \pm 0.03$ & $0.79 \pm 0.04$ \\ \hline
        \textbf{N-IID 0.1} & $0.63 \pm 0.04$ & $0.76 \pm 0.02$ & $0.86 \pm 0.04$  \\ \hline
        & \multicolumn{3}{|c|}{\textbf{Lin attack $(\mathcal{F}^L)$}} \\ \hline
        & \textbf{1 round} & \textbf{3 rounds} & \textbf{5 rounds} \\ \hline
        \textbf{IID} & $0.70 \pm 0.05$ & $0.84 \pm 0.05$ & $0.93 \pm 0.02$ \\ \hline
        \textbf{N-IID 1.0}  & $0.61 \pm 0.06$ & $0.69 \pm 0.03$ & $0.68 \pm 0.04$ \\ \hline
        \textbf{N-IID 0.1} & $0.56 \pm 0.12$ & $0.60 \pm 0.10$ & $0.60 \pm 0.02$ \\ \hline
    \end{tabular}
    \caption{F1-scores of the cosine-based detection. }
    \label{tab:cos_canary}
    \end{subtable}
    \caption{MIA-based average detection performances over $100$ rounds for 1/3/5 canary training epochs. }
    \label{tab:canary}
\end{table}

\subsection{PIA Results}

In our default PIA scenario, we employ $\mathcal{P}^W$ as the attack algorithm. A comparison with $\mathcal{P}^D$ is presented in Table~\ref{tab:PIA_attack}.

\subsubsection{Consistency-based detection}

First, we verify our hypothesis that label distributions for FR clients differ from those of honest clients. In Figure~\ref{fig:div_value}, we present how the inferred label distribution changes round-by-round for an honest (green) and FR client (red). As expected, there is a clear difference between both distributions. Also, for both FR attacks, the FR's consistency score is generally lower than that of honest clients.

\subsubsection{Diversity-based detection}

\begin{figure}[tb]
     \centering
     \includegraphics[width=0.45\textwidth]{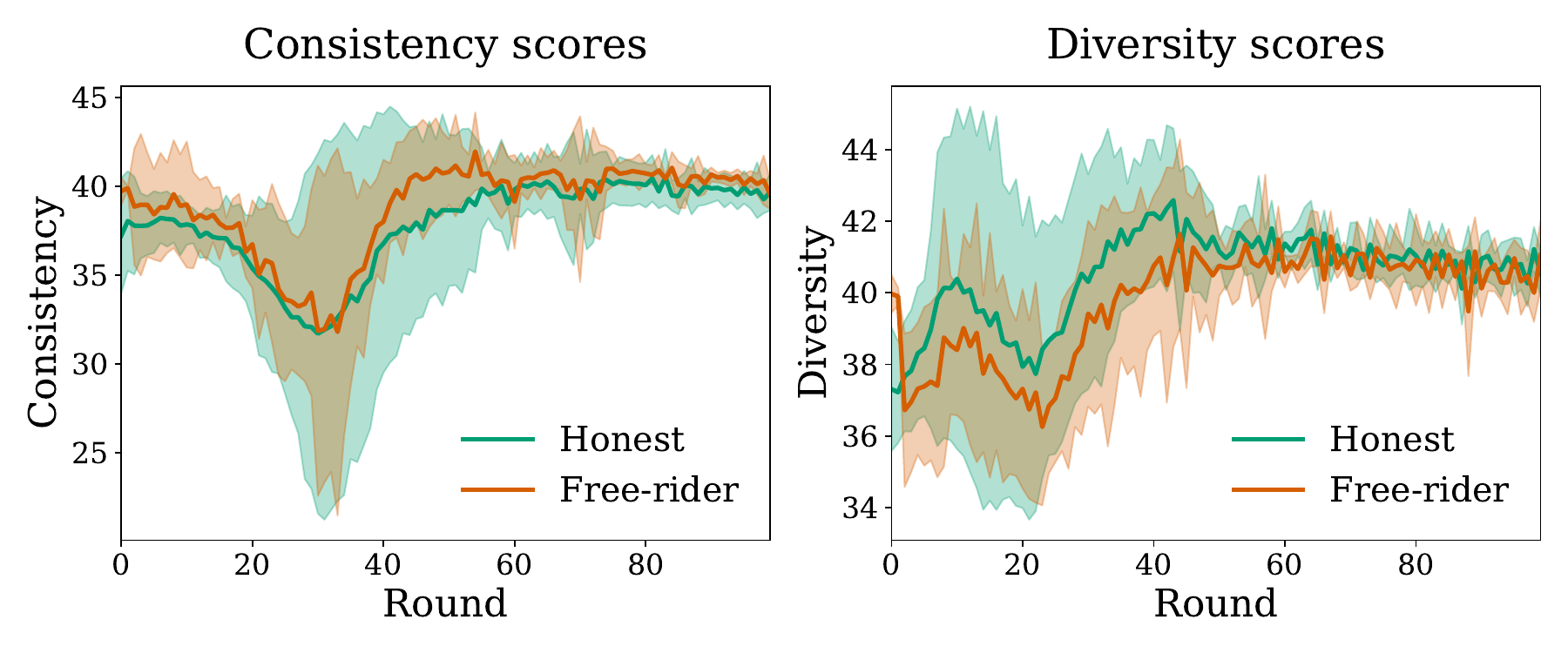}
     \caption{PIA-detection values with $\mathcal{F}^L$. Round-by-round comparison between honest clients (green) and FR clients (red) for consistency (left) and diversity (right).}
     \label{fig:div_value}
\end{figure}

Figure~\ref{fig:div_value} reports the diversity metric for both FR clients and honest clients. The results empirically support our hypothesis that the FR updates can be reconstructed more accurately from a combination of the global model’s label distribution and a uniform distribution than updates from honest clients. As expected, the errors for honest clients (green) and FR clients (red) show a clear separation. 

\subsubsection{Utilized PIA}

\begin{table}[tb]
\footnotesize

    \centering
    \resizebox{0.45\textwidth}{!}{
    \begin{tabular}{|c|c|c|c|c|}
        \hline
        & \multicolumn{4}{|c|}{\textbf{Consistency}} \\ \hline
        & \multicolumn{2}{|c|}{\textbf{Fraboni attack $(\mathcal{F}^F)$}} & \multicolumn{2}{|c|}{\textbf{Lin attack $(\mathcal{F}^L)$}} \\ \hline
        & $\mathcal{P}^W$ & $\mathcal{P}^D$ & $\mathcal{P}^W$ & $\mathcal{P}^D$ \\ \hline
        \textbf{IID}  & $0.99 \pm 0.00$ & $0.71 \pm 0.05$ & $0.84 \pm 0.09$ & $0.75 \pm 0.05$ \\ \hline
        \textbf{N-IID 1.0} & $0.99 \pm 0.00$ & $0.50 \pm 0.12$ & $0.73 \pm 0.09$ & $0.99 \pm 0.00$ \\ \hline
        \textbf{N-IID 0.1} & $0.96 \pm 0.02$ & $0.42 \pm 0.09$ & $0.82 \pm 0.09$ & $0.99 \pm 0.00$ \\ \hline
        & \multicolumn{4}{|c|}{\textbf{Diversity}} \\ \hline
        & \multicolumn{2}{|c|}{\textbf{Fraboni attack $(\mathcal{F}^F)$}} & \multicolumn{2}{|c|}{\textbf{Lin attack $(\mathcal{F}^L)$}} \\ \hline
        & $\mathcal{P}^W$ & $\mathcal{P}^D$ & $\mathcal{P}^W$ & $\mathcal{P}^D$ \\ \hline
        \textbf{IID}  & $0.97 \pm 0.02$ & $0.64 \pm 0.04$ & $0.78 \pm 0.02$ & $0.32 \pm 0.04$ \\ \hline
        \textbf{N-IID 1.0} & $1.00 \pm 0.00$ & $0.56 \pm 0.03$ & $0.84 \pm 0.02$ & $0.35 \pm 0.02$ \\ \hline 
        \textbf{N-IID 0.1} & $1.00 \pm 0.00$ & $0.73 \pm 0.05$ & $0.96 \pm 0.01$ & $0.30 \pm 0.03$ \\ \hline
    \end{tabular}}
    \caption{PIA-based average detection performances over $100$ rounds when either $\mathcal{P}^W$ or $\mathcal{P}^D$ is utilized. }
    \label{tab:PIA_attack}
\end{table}

% \begin{figure*}[tb]
%     \centering
%     \begin{subfigure}{0.45\textwidth}
%     \centering
%     \includegraphics[width=\textwidth]{multiround_mia_yeom.pdf}
%     \caption{Multiround detection performance using loss MIA. } 
%     \label{fig:MIA_loss_multiround}
%     \end{subfigure}
%     \begin{subfigure}{0.45\textwidth}
%     \centering
%     \includegraphics[width=\textwidth]{multiround_mia_cosine.pdf}
%     \caption{Multiround detection performance using cosine MIA.} 
%     \label{fig:MIA_cos_multiround}
%     \end{subfigure}
%     \begin{subfigure}{0.45\textwidth}
%     \centering
%     \includegraphics[width=\textwidth]{multiround_mia_inconsistency.pdf}
%     \caption{Multiround detection performance using consistency PIA.} 
%     \label{fig:PIA_con_multiround}
%     \end{subfigure}
%     \begin{subfigure}{0.45\textwidth}
%     \centering
%     \includegraphics[width=\textwidth]{multiround_pia_dist_score.pdf}
%     \caption{Multiround detection performance using diversity PIA. } 
%     \label{fig:PIA_div_multiround}
%     \end{subfigure}
%     \label{fig:global_comparison}
%     \caption{F1-scores of the proposed FR detection mechanisms. AlexNet is trained on CIFAR-100, distributed across eight clients, where one client is FR.}
% \end{figure*}

In Table~\ref{tab:PIA_attack}, we compare the detection performance of $\mathcal{P}^W$ and $\mathcal{P}^D$ for the two FR strategies, as well as for consistency-based and diversity-based detection mechanisms. We found that while the $\mathcal{P}^W$-based FR detection generally outperforms the $\mathcal{P}^D$-based, it is not the case for consistency-based detection with $\mathcal{F}^L$ as FR strategy. As our goal is not to study the nuanced differences between these attacks, we leave exploring the reason for this as future work. 

% \subsubsection{Multi-round detection}

% Similarly to MIA-based detection, the performance of PIA-based FR detection improves when knowledge is aggregated across multiple rounds. Figures~\ref{fig:PIA_con_multiround} and~\ref{fig:PIA_div_multiround} show the detection rates when decisions from 10 consecutive rounds are utilized. It is important to note that multi-round detection assumes non-adaptive FR attacks, meaning the FR client maintains a consistent strategy across all rounds. For this reason, we refrained from using multi-round detection in all other figures and tables to demonstrate FRIDA's single-round detection effectiveness in scenarios where clients might act as free-riders only in specific rounds.

\subsection{Comparison}

We compare our four proposed methods against three feature-based detection mechanisms in Table~\ref{tab:detection_accuracy} and Table~\ref{tab:ablation}. Overall, FRIDA outperforms or matches the performance of feature-based detection mechanisms across nearly all scenarios, highlighting the positive effect of directly inferring FR properties. Notably, feature-based techniques struggle in IID scenarios and complex datasets such as CIFAR-100, where FRIDA shows superior performance. 

\begin{figure}[tb]
    \centering
    \includegraphics[width=0.5\textwidth]{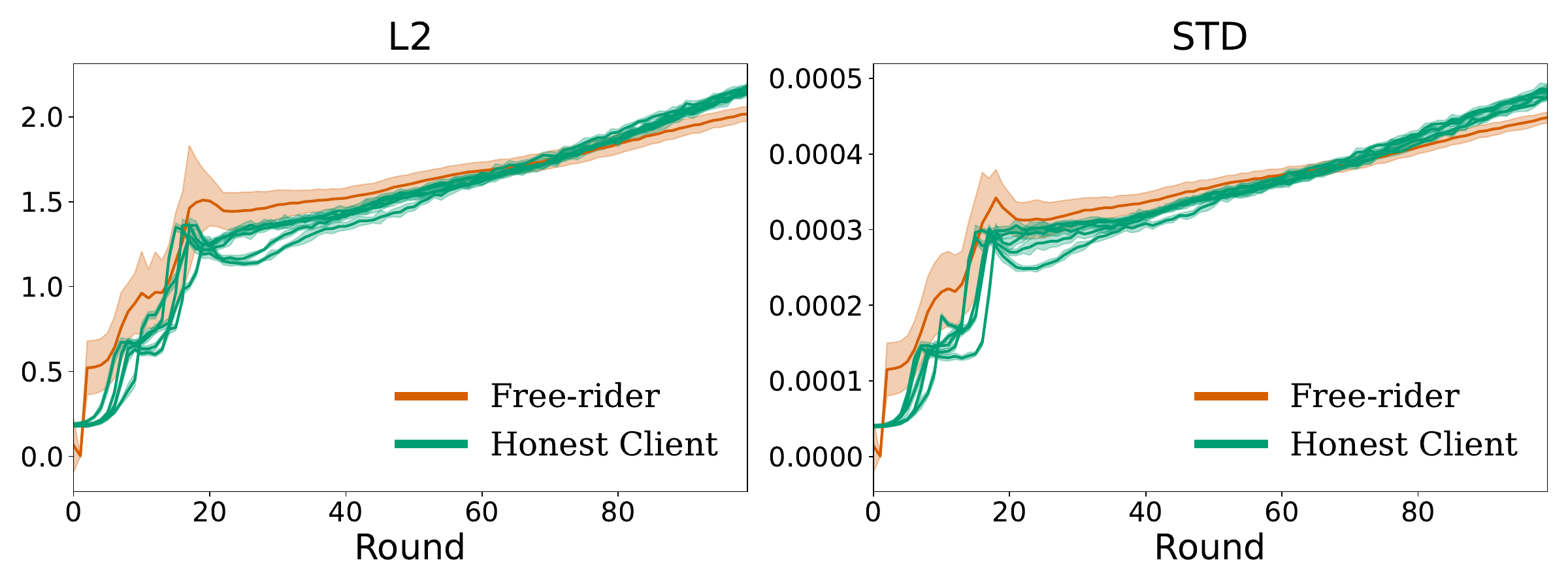}
    \caption{Round-wise comparison of the L2 and STD metrics for honest and FR clients in CIFAR-100. $\mathcal{F}^Z$ successfully mimics the L2 and STD patterns of honest clients. }
    \label{fig:l2_std}
\end{figure}

Figure~\ref{fig:l2_std} shows how $\mathcal{F}^Z$ FR strategy captures geometric properties characteristic of honest clients. In this context, we argue that feature-based detection perpetuates the cat-and-mouse dynamic between attacks and defenses, whereas FRIDA directly targets the underlying indicators of free-riding. This focus enables more robust and fairness-oriented client assessment in federated learning. 

Interestingly, PIA-based detection achieves better results with larger label sets, despite the underlying PIA attack performing worse in these cases. This suggests that less accurate label reconstruction across a wider range of labels provides more valuable information for detection than more precise reconstruction over a narrower label set. 

Finally, while we evaluated all detection mechanisms\textemdash both direct and indirect\textemdash individually in this study, they are complementary and could be integrated to achieve even greater detection performance. 

\subsection{Ablation Study}
\label{sec:ablation}

Here, we conduct a systematic study of how different factors influence FRIDA's detection performance. In addition, we evaluate the impact of privacy-preserving settings by incorporating DP into the local training process. A summary of these results is provided in Table~\ref{tab:ablation}.

\begin{table*}[tb]
\footnotesize	
    \centering
    \begin{tabular}{|c|c|c|c|c|c|c|c|c|}
    \hline
    \multicolumn{9}{|c|}{\textbf{CIFAR-10}} \\ \hline
    \multicolumn{2}{|c|}{} & \multicolumn{2}{|c|}{\textbf{MIA}} & \multicolumn{2}{|c|}{\textbf{PIA}} & \multicolumn{3}{|c|}{\textbf{Feature-based}} \\ \hline
    \multicolumn{2}{|c|}{} & \textbf{Loss} & \textbf{Cosine} & \textbf{Consistency} & \textbf{Divergence} & \textbf{L2} & \textbf{STD} & \textbf{Cosim} \\ \hline

    \multirow{3}{*}{\textbf{1 FR clients}} & \textbf{IID} & $\mathbf{0.94 \pm 0.04}$ & $0.86 \pm 0.14$ & $0.90 \pm 0.13$ & $0.84 \pm 0.15$ & $0.86 \pm 0.09$ & $0.87 \pm 0.10$ & $0.46 \pm 0.22$ \\ \cline{2-9}
    & \textbf{N-IID 1.0} & $0.68 \pm 0.13$ & $0.73 \pm 0.16$ & $0.73 \pm 0.007$ & $0.78 \pm 0.11$ & $\mathbf{0.83 \pm 0.11}$ & $\mathbf{0.83 \pm 0.11}$ & $0.20 \pm 0.09$ \\ \cline{2-9}
    & \textbf{N-IID 0.1} & $0.30 \pm 0.20$ & $0.34 \pm 0.33$ & $0.20 \pm 0.08$ & $\mathbf{0.66 \pm 0.10}$ & $0.56 \pm 0.14$ & $0.56 \pm 0.14$ & $0.19 \pm 0.12$ \\ \hline

    \multirow{3}{*}{\textbf{2 FR clients}} & \textbf{IID} & $0.94 \pm 0.04$ & $0.86 \pm 0.11$ & $\mathbf{0.95 \pm 0.03}$ & $0.87 \pm 0.10$ & $0.86 \pm 0.07$ & $0.87 \pm 0.09$ & $0.59 \pm 0.14$ \\ \cline{2-9}
    & \textbf{N-IID 1.0} & $0.69 \pm 0.11$ & $0.78 \pm 0.10$ & $0.80 \pm 0.00$ & $\mathbf{0.84 \pm 0.06}$ & $0.83 \pm 0.08$ & $0.83 \pm 0.07$ & $0.30 \pm 0.12$ \\ \cline{2-9}
    & \textbf{N-IID 0.1} & $0.34 \pm 0.17$ & $0.21 \pm 0.19$ & $0.25 \pm 0.05$ & $\mathbf{0.65 \pm 0.07}$ & $\mathbf{0.65 \pm 0.08}$ & $\mathbf{0.65 \pm 0.06}$ & $0.20 \pm 0.12$ \\ \hline

    \multicolumn{9}{|c|}{\textbf{CIFAR-100}} \\ \hline

    \multirow{3}{*}{\textbf{1 FR clients}} & \textbf{IID} & $0.92 \pm 0.07$ & $0.96 \pm 0.09$ & $0.97 \pm 0.04$ & $\mathbf{0.98 \pm 0.02}$ & $0.60 \pm 0.2$ & $0.60 \pm 0.20$ & $0.43 \pm 0.23$ \\ \cline{2-9}
    & \textbf{N-IID 1.0} & $0.70 \pm 0.14$ & $0.74 \pm 0.10$ & $0.99 \pm 0.01$ & $\mathbf{1.00 \pm 0.00}$ & $0.83 \pm 0.12$ & $0.83 \pm 0.13$ & $0.17 \pm 0.19$ \\ \cline{2-9}
    & \textbf{N-IID 0.1} & $0.46 \pm 0.19$ & $0.11 \pm 0.67$ & $0.66 \pm 0.09$ & $\mathbf{1.00 \pm 0.00}$ & $0.92 \pm 0.03$ & $0.83 \pm 0.10$ & $0.30 \pm 0.22$ \\ \hline

    \multirow{3}{*}{\textbf{2 FR clients}} & \textbf{IID} & $0.93 \pm 0.06$ & $0.97 \pm 0.03$ & $0.99 \pm 0.02$ & $\mathbf{1.00 \pm 0.00}$ & $0.64 \pm 0.18$ & $0.59 \pm 0.17$ & $0.56 \pm 0.23$ \\ \cline{2-9}
    & \textbf{N-IID 1.0} & $0.71 \pm 0.01$ & $0.75 \pm 0.08$ & $0.99 \pm 0.00$ & $\mathbf{1.00 \pm 0.00}$ & $0.91 \pm 0.08$ & $0.90 \pm 0.07$ & $0.39 \pm 0.00$ \\ \cline{2-9}
    & \textbf{N-IID 0.1} & $0.48 \pm 0.16$ & $0.73 \pm 0.17$ & $0.66 \pm 0.02$ & $\mathbf{1.00 \pm 0.00}$ & $0.86 \pm 0.03$ & $0.84 \pm 0.03$ & $0.66 \pm 0.00$ \\ \hline
    \end{tabular}
    \caption{Average round-by-round F1-scores of the proposed detection techniques and feature-based metrics. AlexNet is trained \textbf{\textit{in a cross-silo scenario}}with 8 clients, which one or two are FR using $\mathcal{F}^Z$.}
    \label{tab:detection_accuracy}
\end{table*}

\subsubsection{Dataset \& Model}

We evaluate the methods on a more complex dataset using CIFAR-100 and a deeper architecture (VGG-19). In this scenario, we observe that most detection mechanisms improve the performance. We attribute this improvement to the fact that the global model has not yet fully converged, allowing FR updates to blend easily with the updates of honest clients, since differences induced from local heterogeneous datasets have not yet emerged. For PIA in particular, increasing the number of labels enhances detection performance, as it provides more data points for its output, thereby enhancing its ability to discriminate between honest and FR clients.
%
% \subsubsection{FR Ratio}

% Table~\ref{tab:detection_accuracy} and Table~\ref{tab:ablation} present the results for 8 clients with 2 and 4 FR clients. As the number of FR clients increases, all anomaly-based FR detection methods are affected by the inherent limitations of the $z$-score outlier detection mechanism. However, for $50\%$ of FR, feature-based metrics have a poor F1-score while FRIDA succeeds, demonstrating improved robustness when directly inferring FR properties.
% TODO table 5 anf 6 not contains 50% case

\subsubsection{FR Strategy}

By comparing the detection performance reported in Table~\ref{tab:detection_accuracy} (using $\mathcal{F}^Z$) and Table~\ref{tab:ablation} in the Appendix (using $\mathcal{F}^F$ and $\mathcal{F}^L$), we observe that $\mathcal{F}^Z$ is the most stealthy among the evaluated free-rider strategies. This attack explicitly aligns the free-rider's updates with the geometric characteristics of honest clients, making it significantly harder to detect, especially in IID scenarios, where client updates tend to be more similar. In contrast, $\mathcal{F}^F$ and $\mathcal{F}^L$ are simpler strategies that do not attempt to replicate the geometric structure of honest updates. As a result, they are more easily identified by feature-based detection methods, particularly in non-IID settings where honest updates naturally diverge due to the heterogeneity of the local datasets.

\begin{figure}[tb]
    \centering
    \begin{subfigure}{0.45\textwidth}
    \centering
    \includegraphics[width=0.7\textwidth]{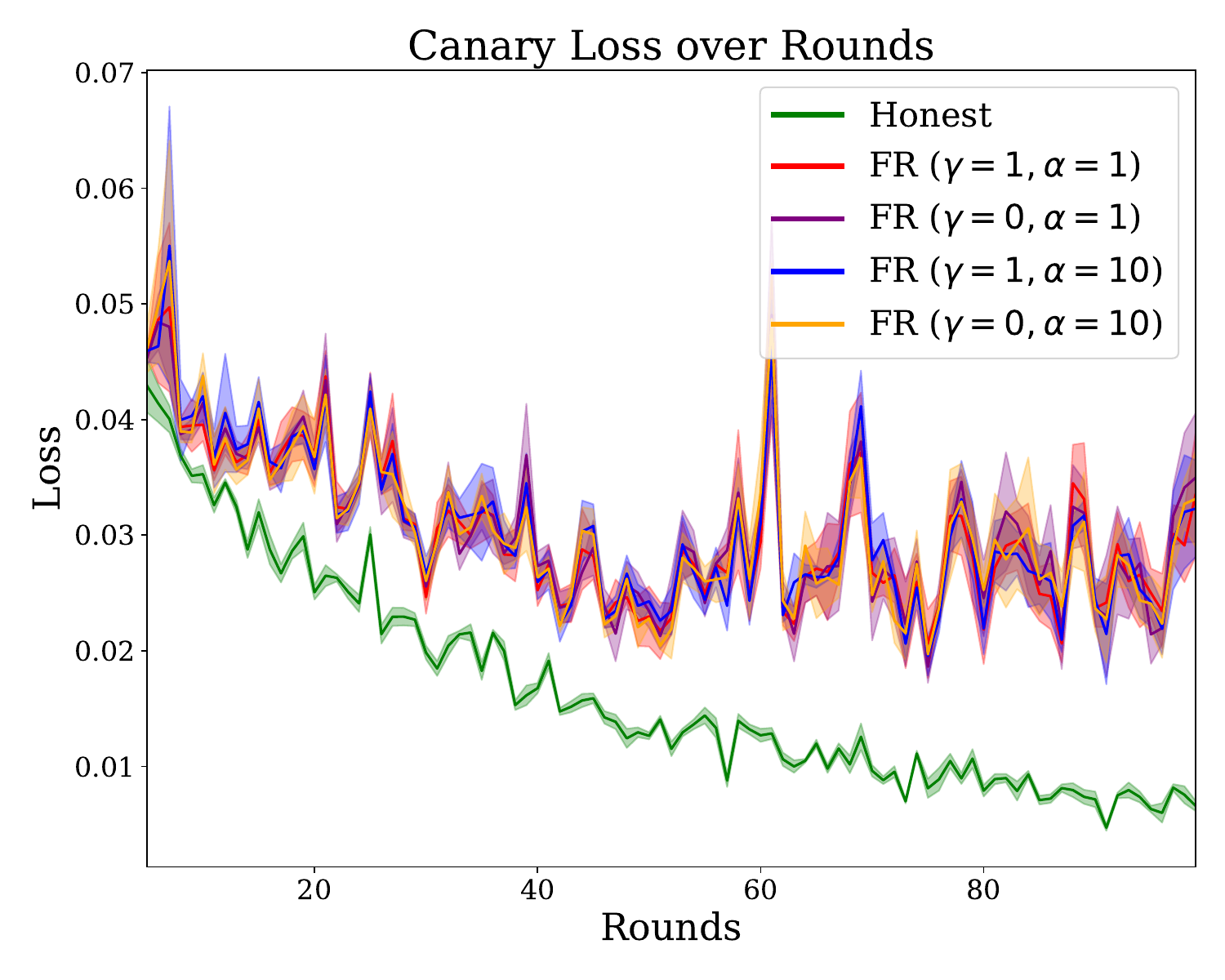}
    \caption{Round-wise canary losses for different noise configurations. We observe how the free-rider loss remains constant across all configurations.}
    \label{fig:ablation_canary}
    \end{subfigure}
    \begin{subfigure}{0.45\textwidth}
    \centering
    \includegraphics[width=\textwidth]{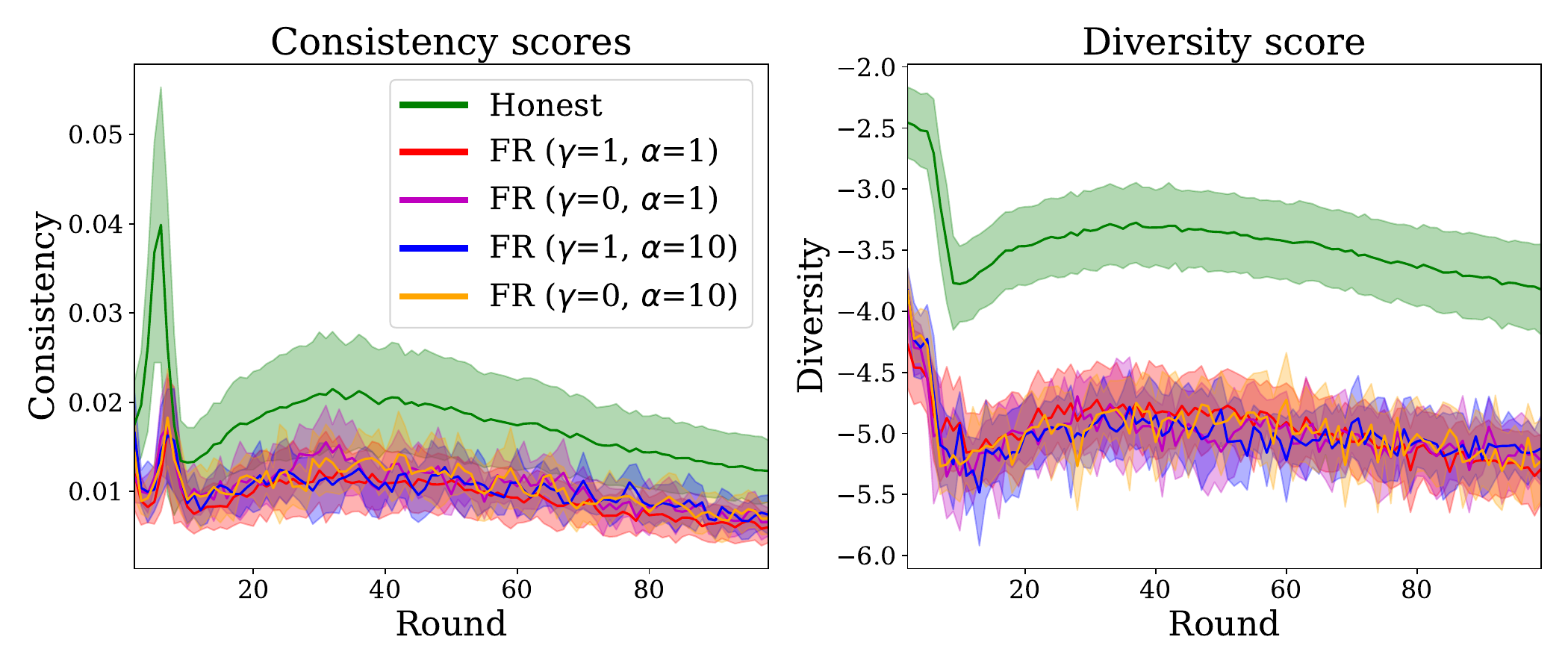}
    \caption{Round-wise consistency values (left) and diversity values (right) for different noise configurations. In both cases, the free-rider scores remain constant.}
    \label{fig:ablation_pia}
    \end{subfigure}
    \caption{Comparison of detection metrics across various noise configurations for PIA and MIA detection methods.}
    \label{fig:ablation_noise}
\end{figure}

\subsubsection{FR Noise}

We complement our analysis by studying different noise rates for $\mathcal{F}^L$. In this regard, we vary the noise multiplier $\alpha$ and the noise power $\gamma$, which is the exponent applied to the iteration number (see Equation~\ref{eq:Lin} in the Appendix). These parameters affect the standard deviation of the Gaussian distribution $\mathcal{N}(G^{t-1}, \alpha \cdot \sigma \cdot t^{\gamma})$ from which the FR client draws the noise. As shown in Figure~\ref{fig:ablation_canary} and Figure~\ref{fig:ablation_canary}, changing $\alpha$ and $\gamma$ has minimal impact on FRIDA's detection performance. The canary loss, as well as the consistency and diversity scores, remain largely unaffected, indicating that FRIDA is robust to variations in noise intensity within this class of FR strategy.

\subsubsection{Clients}
\label{ablation:num_clients}

We evaluate the methods in a cross-edge setting with a larger number of clients and a smaller 500k parameter CNN. Table~\ref{tab:100clients} reports results for 100 clients, with 10, 20, and 40 FR clients. We observe that FRIDA consistently outperforms feature-based methods across all settings. Interestingly, the cosine-based detector shows excellent performance even with a large number of free-riders, since it infers membership per client and does not rely on cohort statistics, offering a promising alternative to detectors that depend on outlier rules. Because we standardize all decision rules to a $z$-score policy, scenarios with more free-riders than honest clients are excluded, as under such conditions honest clients would become outliers. In this regime, one must adopt an absolute threshold policy \cite{wang2023frad}, which is orthogonal to our evaluation, as our focus lies on comparing detection metrics rather than tuning decision thresholds. 
% This limitation particularly impacts STD-DAGMM, for which prior work \cite{wang2023frad} reported higher performance by fine-tuning thresholds on the training data.

\begin{figure}[tb]
    \centering
    \includegraphics[width=0.5\textwidth]{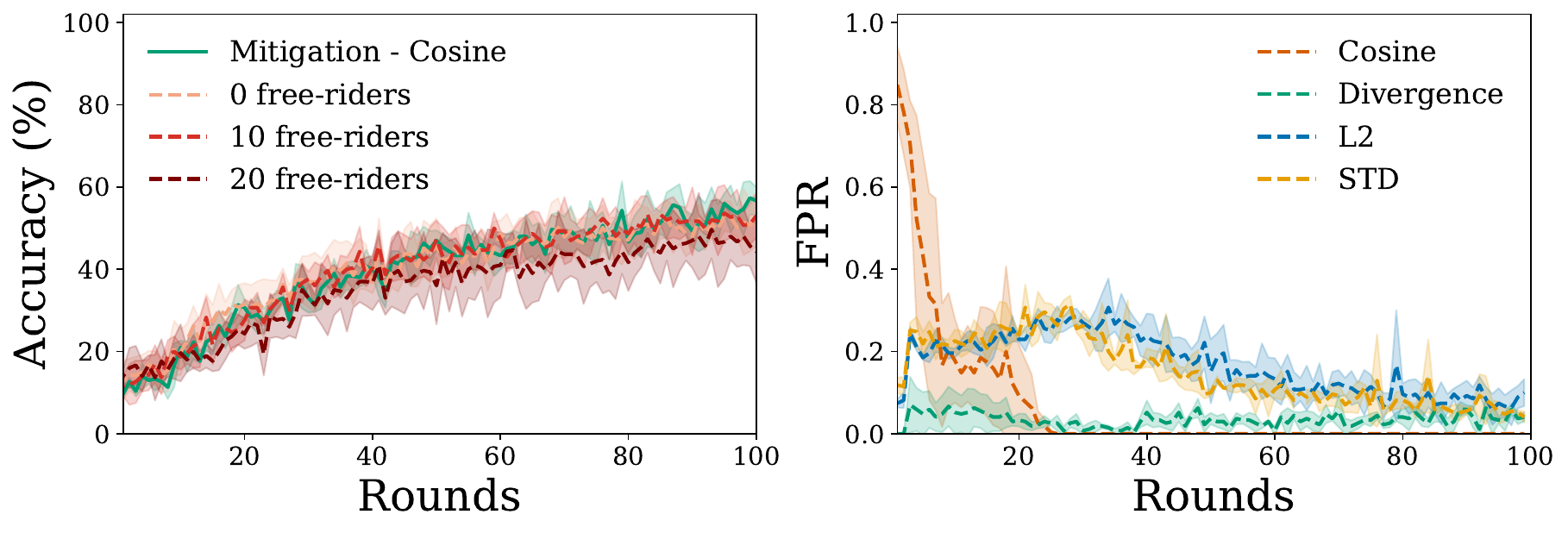}
    \caption{(\textit{Left}) Global model accuracy over training rounds for 100 clients with $0\%$, $10\%$, and $20\%$ free-riders, and with cosine-based detection removing detected free-riders from aggregation. (\textit{Right}) Per-round False Positive Rate (FPR) comparison with $10\%$ free-riders (100 clients) across detection methods.}
    \label{fig:auc}
\end{figure}

\begin{figure}[tb]
    \centering
    \includegraphics[width=0.5\textwidth]{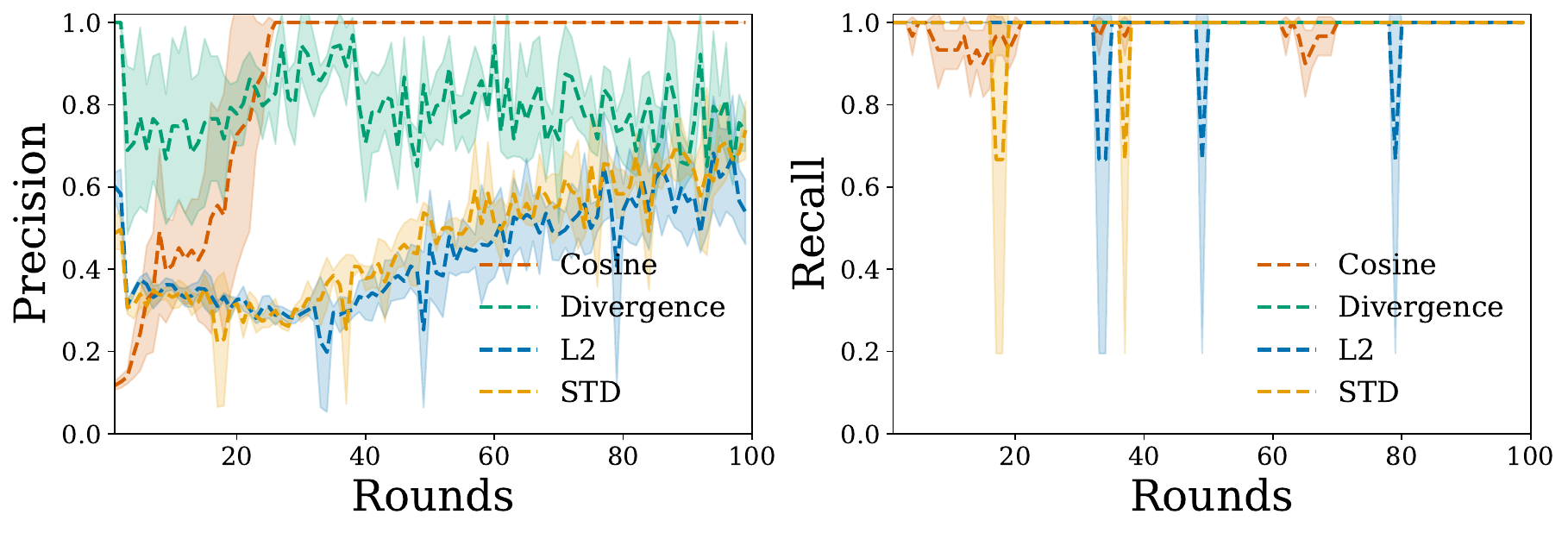}
    \caption{Per-round precision (left) and recall (right) for $10\%$ free-riders in a 100-client IID FL setup. Divergence-based and cosine-based detection achieve higher precision than L2 and STD feature-based methods, leading to lower false positive rates.}
    \label{fig:prec_rec}
\end{figure}

Moreover, we compare FPR, precision, and recall across detection methods in Figures~\ref{fig:auc} and~\ref{fig:prec_rec}. Overall, the cosine and divergence FRIDA methods, representing the best-performing MIA and PIA approaches, maintain lower FPRs than feature-based detectors, which is critical in security-sensitive FL to avoid penalizing honest clients. Figure~\ref{fig:auc} (left) further shows the effect of removing free-riders using cosine-based detection, with the global model converging with a pattern closely matching the 0~free-rider scenario.

\begin{table}[tbh]
    \footnotesize	
    \centering
    \begin{subtable}{0.5\textwidth}
    \centering
    \begin{tabular}{|c|c|c|c|}
        \hline
        \multicolumn{4}{|c|}{\textbf{10 FR clients - 100 clients}} \\ \hline
        & \textbf{Fraboni $(\mathcal{F}^F)$} & \textbf{Lin $(\mathcal{F}^L)$} & \textbf{Zhu $(\mathcal{F}^Z)$} \\ \hline
        \textbf{Loss} & $0.26 \pm 0.06$ & $0.56 \pm 0.07$ & $0.52 \pm 0.12$ \\ \hline
        \textbf{Cosine} & $\mathbf{0.91 \pm 0.04}$ & $\mathbf{0.94 \pm 0.02}$ & $\mathbf{0.93 \pm 0.03}$ \\ \hline
        \textbf{Consistency} & $0.76 \pm 0.09$ & $0.74 \pm 0.13$ & $0.71 \pm 0.11$ \\ \hline
        \textbf{Divergence} & $0.79 \pm 0.13$ & $0.78 \pm 0.08$ & $0.76 \pm 0.073$ \\ \hline
        \textbf{L2} & $0.80 \pm 0.06$ & $0.62 \pm 0.05$ & $0.60 \pm 0.06$ \\ \hline
        \textbf{STD} & $0.84 \pm 0.06$ & $0.63 \pm 0.06$ & $0.56 \pm 0.04$ \\ \hline
        \textbf{Cosim} & $0.22 \pm 0.14 $ & $0.09 \pm 0.10$ & $0.08 \pm 0.12$ \\ \hline
        \textbf{STD-DAGMM} & $0.63 \pm 0.14 $ & $0.29 \pm 0.15$ & $0.71 \pm 0.15$ \\ \hline
    \end{tabular}
    \caption{F1-scores of the proposed detection techniques for CNN with CIFAR-10 where 10 out of 100 IID clients are FR. }
    \label{tab:100clients_10fr}
    \end{subtable}
    \vspace{0.1cm}
    \begin{subtable}{0.5\textwidth}
    \centering
    \begin{tabular}{|c|c|c|c|}
        \hline
        \multicolumn{4}{|c|}{\textbf{20 FR clients - 100 clients}} \\ \hline
        & \textbf{Fraboni $(\mathcal{F}^F)$} & \textbf{Lin $(\mathcal{F}^L)$} & \textbf{Zhu $(\mathcal{F}^Z)$} \\ \hline
        \textbf{Loss} & $0.35 \pm 0.10$ & $0.66 \pm 0.08$ & $0.65 \pm 0.01$ \\ \hline
        \textbf{Cosine} & $\mathbf{0.90 \pm 0.06}$ & $\mathbf{0.94 \pm 0.01}$ & $\mathbf{0.94 \pm 0.03}$ \\ \hline
        \textbf{Consistency} & $0.82 \pm 0.12$ & $0.88 \pm 0.07$ & $0.86 \pm 0.05$ \\ \hline
        \textbf{Divergence} & $0.90 \pm 0.04$ & $0.88 \pm 0.06$ & $0.85 \pm 0.10$ \\ \hline
        \textbf{L2} & $0.87 \pm 0.00$ & $0.76 \pm 0.04$ & $0.72 \pm 0.00$ \\ \hline
        \textbf{STD} & $0.90 \pm 0.04$ & $0.49 \pm 0.14$ & $0.81 \pm 0.03$ \\ \hline
        \textbf{Cosim} & $0.34 \pm 0.17 $ & $0.12 \pm 0.14$ & $0.11 \pm 0.13$ \\ \hline
        \textbf{STD-DAGMM} & $0.42 \pm 0.31 $ & $0.24 \pm 0.00$ & $0.45 \pm 0.33$ \\ \hline
    \end{tabular}
    \caption{F1-scores of the proposed detection techniques for CNN with CIFAR-10 where 20 out of 100 IID clients are FR.}
    \label{tab:100clients_20fr}
    \end{subtable}
    \vspace{0.1cm}
    \begin{subtable}{0.5\textwidth}
    \centering
    \begin{tabular}{|c|c|c|c|}
        \hline
        \multicolumn{4}{|c|}{\textbf{40 FR clients - 100 clients}} \\ \hline
        & \textbf{Fraboni $(\mathcal{F}^F)$} & \textbf{Lin $(\mathcal{F}^L)$} & \textbf{Zhu $(\mathcal{F}^Z)$} \\ \hline
        \textbf{Loss} & $0.33 \pm 0.06$ & $0.48 \pm 0.05$ & $0.47 \pm 0.04$ \\ \hline
        \textbf{Cosine} & $\mathbf{0.95 \pm 0.02}$ & $\mathbf{0.96 \pm 0.02}$ & $\mathbf{0.97 \pm 0.01}$ \\ \hline
        \textbf{Consistency} & $0.85 \pm 0.04$ & $0.59 \pm 0.11$ & $0.85 \pm 0.01$ \\ \hline
        \textbf{Divergence} & $0.67 \pm 0.13$ & $0.62 \pm 0.11$ & $0.81 \pm 0.03$ \\ \hline
        \textbf{L2} & $0.86 \pm 0.02$ & $0.55 \pm 0.08$ & $0.73 \pm 0.13$ \\ \hline
        \textbf{STD} & $0.86 \pm 0.02$ & $0.33 \pm 0.19$ & $0.72 \pm 0.13$ \\ \hline
        \textbf{Cosim} & $0.28 \pm 0.19$ & $0.16 \pm 0.16$ & $0.37 \pm 0.29$ \\ \hline
        \textbf{STD-DAGMM} & $0.01 \pm 0.00$ & $0.05 \pm 0.02$ & $0.00 \pm 0.00$ \\ \hline
    \end{tabular}
    \caption{F1-scores of the proposed detection techniques for CNN with CIFAR-10 where 40 out of 100 IID clients are FR.}
    \label{tab:100clients_40fr}
    \end{subtable}
    % \vspace{0.1cm}
    % \begin{subtable}{0.5\textwidth}
    % \centering
    % \begin{tabular}{|c|c|c|c|}
    %     \hline
    %     \multicolumn{4}{|c|}{\textbf{60 FR clients - 100 clients}} \\ \hline
    %     & \textbf{Fraboni $(\mathcal{F}^F)$} & \textbf{Lin $(\mathcal{F}^L)$} & \textbf{Zhu $(\mathcal{F}^Z)$} \\ \hline
    %     \textbf{Loss} & -- & -- & $0.338 \pm 0.032$ \\ \hline
    %     \textbf{Cosine} & -- & -- & -- \\ \hline
    %     \textbf{Consistency} & -- & -- & -- \\ \hline
    %     \textbf{Divergence} & -- & -- & -- \\ \hline
    %     \textbf{L2} & -- & -- & -- \\ \hline
    %     \textbf{STD} & -- & -- & -- \\ \hline
    %     \textbf{Cosim} & -- & -- & -- \\ \hline
    %     \textbf{STD-DAGMM} & -- & -- & -- \\ \hline
    % \end{tabular}
    % \caption{F1-scores of the proposed detection techniques for CNN with CIFAR-10 where 60 out of 100 IID clients are FR.}
    % \label{tab:100clients_60fr}
    % \end{subtable}
    
    \caption{Comparison across detection methods on a cross-edge scenario with 100 clients and 10/20/40 FR clients.}
    \label{tab:100clients}
\end{table}

\subsubsection{DP Noise}

We consider the deployment of FRIDA under differential privacy, which adds Gaussian noise to the honest clients to conceal individual updates. Specifically, we applied local Renyi DP~\cite{pejo2022guide} with $\delta=|D_n|^{-1}=1/4000$, set the clipping threshold to $0.5$, the batch size to $32$, the training rounds to $50$, and the noise variance to $\sigma=\{0.7,1.0,1.5\}$. These values correspond to protection levels of $\varepsilon=\{9,4,2\}$, calculated using the momentum accountant technique~\cite{abadi2016deep}. The corresponding results are presented in Table~\ref{tab:DP}. 

As expected, adding DP degrades FRIDA's detection performance, particularly for MIA-based detection, which relies on fine-grained individual-level signals. As noise increases, these signals become increasingly indistinguishable, leading to a drop in F1 score. In contrast, PIA-based detection remains robust under local DP, as it captures broader statistical patterns across the client’s dataset, which are less affected by individual-level noise. In this regard, both diversity and consistency-based detection remain effective in identifying free-riders. These findings indicate that, under local DP, PIA should be the default detection mechanism, while MIA is only applicable in no-noise scenarios. Finally, we note that FRIDA assumes visibility over per-client updates. In setups involving secure aggregation, detection would require an additional mechanism—such as a two-server architecture—to decouple update inspection from client identity while maintaining privacy guarantees.

\begin{table}[tbh]
    \centering
    \resizebox{0.45\textwidth}{!}{
    \begin{tabular}{|c|c|c|c|c|c|}
    \hline
    \multicolumn{2}{|c|}{}& \multicolumn{2}{|c|}{\textbf{MIA}} & \multicolumn{2}{|c|}{\textbf{PIA}} \\ \hline
    $\varepsilon$ & FR & \textbf{Loss} & \textbf{Cosine} & \textbf{Consistency} & \textbf{Divergence} \\ \hline
     \multirow{3}{*}{9} & $\mathcal{F}^F$ & $0.04 \pm 0.08$ & $0.23 \pm 0.05$ & $0.98 \pm 0.02$ & $\mathbf{0.98 \pm 0.01}$ \\ 
    \cline{2-6}
    & $\mathcal{F}^L$ & $0.23 \pm 0.23$ & $0.25 \pm 0.00$ & $\mathbf{0.60 \pm 0.24}$ & $0.30 \pm 0.23$ \\ 
    \cline{2-6}
    & $\mathcal{F}^Z$ & $0.21 \pm 0.12$ & $0.20 \pm 0.03$ & $0.72 \pm 0.14$ & $\mathbf{0.90 \pm 0.12}$ \\ 
    \hline
     \multirow{3}{*}{4} & $\mathcal{F}^F$ & $0.04 \pm 0.08$ & $0.23 \pm 0.00$ & $0.97 \pm 0.03$ & $\mathbf{0.98 \pm 0.01}$ \\ 
    \cline{2-6}
    & $\mathcal{F}^L$ & $0.18 \pm 0.19$ & $0.23 \pm 0.00$ & $\mathbf{0.61 \pm 0.21}$ & $0.16 \pm 0.12$ \\ 
    \cline{2-6}
    & $\mathcal{F}^Z$ & $0.17 \pm 0.21$ & $0.23 \pm 0.01$ & $0.61 \pm 0.14$ & $\mathbf{0.90 \pm 0.13}$ \\ 
    \hline
     \multirow{3}{*}{2} & $\mathcal{F}^F$ & $0.04 \pm 0.08$ & $0.23 \pm 0.00$ & $0.98 \pm 0.02$ & $\mathbf{0.97 \pm 0.00}$ \\ 
    \cline{2-6}
    & $\mathcal{F}^L$ & $0.20 \pm 0.15$ & $0.22 \pm 0.03$ & $\mathbf{0.62 \pm 0.22}$ & $0.39 \pm 0.30$ \\ 
    \cline{2-6}
    & $\mathcal{F}^Z$ & $0.14 \pm 0.18$ & $0.25 \pm 0.01$ & $0.57 \pm 0.18$ & $\mathbf{0.88 \pm 0.15}$ \\ 
    \hline
    \end{tabular}}
    \caption{Average round-by-round F1-scores of the proposed detection techniques and feature-based metrics under an IID scenario, considering varying levels of DP protection applied by honest clients. AlexNet is trained with 8 clients, including one FR.}
    \label{tab:DP}
\end{table}

\section{Discussion}
\label{sec:discussion}
Advanced FR strategies often evade feature-based detection mechanisms, particularly in IID scenarios and complex tasks. With FRIDA, we propose a framework of inference-based mechanisms that directly addresses the fundamental issue underlying FR: the absence of genuine client training contributions. Our approach is FR strategy-agnostic, detecting free-riders even as new FR techniques circumvent existing feature-based defences.

FRIDA's approach presents a direct trade-off by repurposing privacy vulnerabilities to enforce security and fairness. This strategy is highly practical in specific contexts, particularly in systems where implementing strong privacy-preserving techniques imposes prohibitive computational overhead or degrades model utility. Moreover, it is well-suited for systems where the primary objective of guaranteeing fair contributions among clients justifies the need for individual inspection. At the same time, FRIDA must be applied responsibly, with explicit guarantee of client-aware compliance.

Despite this trade-off, FRIDA is not inherently incompatible with privacy. The framework can be deployed within privacy-enhancing architectures, such as two-server models that decouple client identities from their updates. Furthermore, our results demonstrate that PIA-based detection, which relies on dataset-level properties, remains robust even when clients apply local differential privacy to their gradients. This shows that a level of security against free-riding can be maintained without completely abandoning client-side privacy protections.

\subsection{FR detection with MIA}

\begin{figure}[tb]
    \centering
    \includegraphics[width=0.5\textwidth]{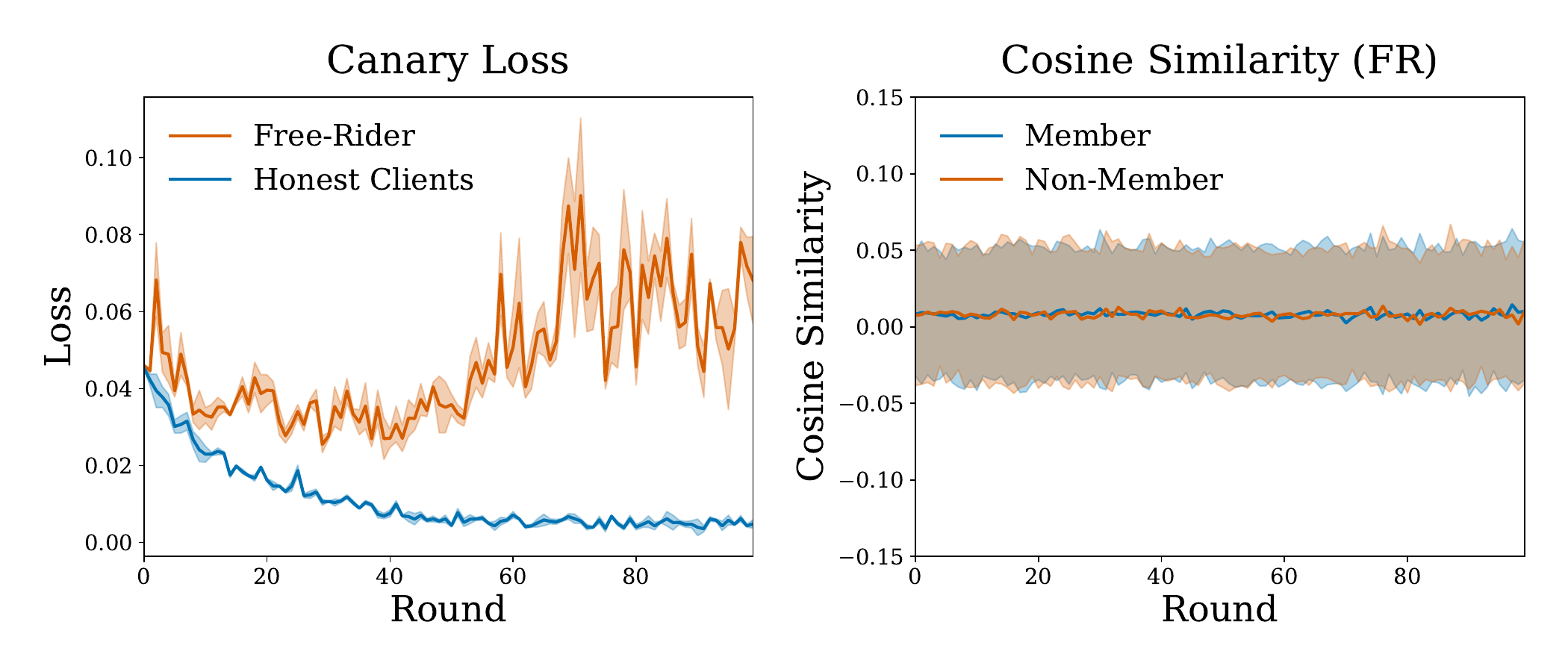}
    \caption{Round-wise comparison of the canary loss and cosine similarity across rounds for an adaptive selfish free-rider which has trained only on the canary dataset. }
    \label{fig:combined_plot}
\end{figure}

While this paper focuses on anonymous free-riders who lack resources, we also tested FRIDA against an adaptive, selfish free-rider who trains on the canary set to evade detection. Our findings reveal a key difference between our MIA methods, shown in Figure \ref{fig:combined_plot}. Loss-based detection remains effective, as the free-rider's canary loss is still distinguishable from that of honest clients. However, the attacker successfully evades the cosine-based method, as the free-rider’s cosine similarity profile becomes statistically identical to an honest client's. This underscores the importance of our PIA-based approaches in these scenarios, which are immune to this adaptive strategy.

Furthermore, this result presents a clear trade-off for the cosine-based method. Its primary advantage is that it evaluates each client in isolation, addressing a key limitation of group-based outlier detection \cite{chen2024rethinking}, which can lose power as the number of free-riders increases. However, it is vulnerable to adaptive selfish attackers. Although we evaluated FRIDA's methods separately, a promising direction for future work is to combine MIA and PIA to create a hybrid defense that is resilient to both anonymous and adaptive FR strategies. For instance, one could apply a sequential strategy, where PIA is used to verify clients only when MIA outputs are ambiguous. However, this combination would come with a higher computational cost. Future work should explore light-weight approximations to privacy attacks, such as using only a subset of layers of the model for the cosine similarity computation, and consider other privacy attacks to improve robustness against potential long-term adaptive free-riders.

\subsection{FR detection with PIA}

Our results indicate that PIA is particularly effective in scenarios with a large number of labels, as it provides more granular information to distinguish between clients. This suggests that the inferred label distribution is a valuable indicator for detecting free-riders, achieving near-perfect F1-scores on complex datasets where feature-based methods fail. Additionally, relying on dataset-level properties makes PIA-based detection robust in DP scenarios where other methods, including MIA, lose effectiveness. Unlike methods requiring client-side training on a canary set, PIA shifts the computational overhead to the server. This introduces an interesting trade-off: while PIA reduces the need for client compliance, it increases the server's computational demands, making it ideal for deployments where client modifications are infeasible or where adaptive selfish attackers are a concern.

\section{Conclusion}
\label{sec:con}

In this work, we introduced FRIDA (\textbf{f}ree-\textbf{ri}der \textbf{d}etection using privacy \textbf{a}ttacks), a framework that leverages techniques from privacy analysis to detect free-riders. We present a new family of detection methods that uncovers synergies between privacy (via inference attacks) and security (via misbehaviour detection) within federated learning. Specifically, we proposed four detection mechanisms, based on membership and property inference attacks, that use local information indicative of real training. We demonstrated the robustness of FRIDA across various scenarios, and how it remains compatible with privacy-preserving techniques. Our approach outperforms feature-based mechanisms that target indirect effects of free-riding.

\subsubsection*{Limitations \& Future Works}

While FRIDA demonstrates strong performance, certain areas remain open for future exploration. We used $z$-score across all detection methods for its simplicity, but other statistical methods could improve the robustness in scenarios with very few or many free-riders. Also, fine-tuning the corresponding detection threshold for each use-case would further increase FRIDA's detection rates. Further future work includes studying the trade-off between exploration and exploitation, specifically determining the optimal round for removing detected clients and its impact on the global model's performance. 

% Our current evaluation focuses on Computer Vision tasks; extending this work to Natural Language Processing (NLP) would further validate its applicability across diverse domains. 

%%% TODO %%%
%Combination of methods
% on-the-fly utilization
%Exploration vs Exploitation trade-off.

\section*{Acknowledgments}
Project no. 145832, implemented with the support provided by the Ministry of Innovation and Technology from the NRDI Fund, financed under the PD\_23 funding scheme.

Also, partially financed by the Spanish Ministry of Science (MICINN), the Research State Agency (AEI), and European Regional Development Funds (ERDF/FEDER) under grant agreement PID2021-126248OB-I00, MCIN/AEI/10.13039/501100011033/FEDER, UE, and Generalitat de Catalunya (AGAUR) under grant 2021-SGR-00478.

Lastly, also funded by the European Union (Grant Agreement Nr. 101095717, SECURED Project). Views and opinions expressed are those of the author(s) only and do not necessarily reflect those of the European Union or the granting authority.

\bibliographystyle{elsarticle-num}
\bibliography{main}

\appendix
\section{Free-riding Attacks}
\label{app:attacks}

\begin{table*}[tb]
    \footnotesize
    \centering
    \begin{tabular}{|c|c|c|c|c|c|c|c|c|}
    \hline
    & & \multicolumn{2}{|c|}{\textbf{MIA}} & \multicolumn{2}{|c|}{\textbf{PIA}} & \multicolumn{3}{|c|}{\textbf{Feature-based}} \\ \hline
    & & \textbf{Loss} & \textbf{Cosine} & \textbf{Consistency} & \textbf{Divergence} & \textbf{L2} & \textbf{STD} & \textbf{Cosim} \\ \hline

    \multirow{3}{*}{\textbf{A-10-8-L}}  & \textbf{IID} & $0.89 \pm 0.02$ & $0.71 \pm 0.04$ & $0.84 \pm 0.06$ & $0.78 \pm 0.03$ & $\mathbf{0.98 \pm 0.01}$ & $\mathbf{0.98 \pm 0.01}$ & $0.42 \pm 0.24$ \\ \cline{2-9}
    & \textbf{N-IID 1.0} & $0.59 \pm 0.02$ & $0.59 \pm 0.05$ & $0.66 \pm 0.00$ & $0.70 \pm 0.01$ & $\mathbf{0.91 \pm 0.01}$ & $\mathbf{0.91 \pm 0.01}$ & $0.22 \pm 0.13$ \\ \cline{2-9}
    & \textbf{N-IID 0.1} & $0.27 \pm 0.03$ & $0.52 \pm 0.13$ & $0.05 \pm 0.03$ & $\mathbf{0.61 \pm 0.02}$ & $0.50 \pm 0.01$ & $0.49 \pm 0.01$ & $ 0.16 \pm 0.10$ \\ \hline

    \multirow{3}{*}{\textbf{A-100-8-L}}  & \textbf{IID} & $0.87 \pm 0.02$ & $0.94 \pm 0.03$ & $0.95 \pm 0.02$ & $\mathbf{1.00 \pm 0.00}$ & $\mathbf{1.00 \pm 0.01}$ & $0.99 \pm 0.01$ & $0.34 \pm 0.24$\\ \cline{2-9}
    & \textbf{N-IID 1.0} & $0.61 \pm 0.02$ & $0.60 \pm 0.03$ & $0.97 \pm 0.02$ & $\mathbf{1.00 \pm 0.00}$ & $\mathbf{1.00 \pm 0.00}$ & $\mathbf{1.00 \pm 0.00}$ & $0.19 \pm 0.22$ \\ \cline{2-9}
    & \textbf{N-IID 0.1} & $0.41 \pm 0.04$ & $0.74 \pm 0.07$ & $0.44 \pm 0.05$ & $\mathbf{1.00 \pm 0.00}$ & $\mathbf{1.00 \pm 0.00}$ & $\mathbf{1.00 \pm 0.00}$ & $0.36 \pm 0.23$ \\ \hline

    \multirow{3}{*}{\textbf{A-10-8-F}} & \textbf{IID} & $0.75 \pm 0.02$ & $0.74 \pm 0.03$ & $0.97 \pm 0.01$ & $0.97 \pm 0.01$ & $\mathbf{1.00 \pm 0.00}$ & $\mathbf{1.00 \pm 0.00}$ & $0.56 \pm 0.20$ \\ \cline{2-9}
    & \textbf{N-IID 1.0} & $0.41 \pm 0.01$ & $0.61 \pm 0.05$ & $0.67 \pm 0.01$ & $0.91 \pm 0.02$ & $\mathbf{0.98 \pm 0.00}$ & $\mathbf{0.98 \pm 0.01}$ & $0.08 \pm 0.05$ \\ \cline{2-9}
    & \textbf{N-IID 0.1} & $0.40 \pm 0.01$ & $0.75 \pm 0.08$ & $0.66 \pm 0.01$ & $\mathbf{1.00 \pm 0.00}$ & $0.58 \pm 0.01$ & $0.58 \pm 0.01$ & $0.01 \pm 0.01$ \\ \hline

    \multirow{3}{*}{\textbf{A-100-8-F}} & \textbf{IID}  & $0.73 \pm 0.01$ & $0.92 \pm 0.02$ & $0.99 \pm 0.01$ & $\mathbf{1.00 \pm 0.00}$ & $\mathbf{1.00 \pm 0.00}$ & $\mathbf{1.00 \pm 0.00}$ & $0.64 \pm 0.20$ \\ \cline{2-9}
    & \textbf{N-IID 1.0} & $0.40 \pm 0.02$ & $0.58 \pm 0.03$ & $0.99 \pm 0.01$ & $\mathbf{1.00 \pm 0.00}$ & $\mathbf{1.00 \pm 0.00}$ & $\mathbf{1.00 \pm 0.00}$ & $0.10 \pm 0.12$ \\ \cline{2-9}
    & \textbf{N-IID 0.1} & $0.51 \pm 0.04$ & $0.87 \pm 0.06$ & $0.99 \pm 0.01$ & $\mathbf{1.00 \pm 0.00}$ & $\mathbf{1.00 \pm 0.00}$ & $\mathbf{1.00 \pm 0.00}$ & $0.22 \pm 0.21$\\ \hline
    
    \multirow{3}{*}{\textbf{V-10-4-F}} & \textbf{IID} & $0.85 \pm 0.02$ & $0.82 \pm 0.01$ & $\mathbf{0.99 \pm 0.01}$ & $\mathbf{0.99 \pm 0.01}$ & $0.98 \pm 0.01$ & $0.98 \pm 0.01$ & $0.68 \pm 0.34$ \\ \cline{2-9}
    & \textbf{N-IID 1.0} & $0.70 \pm 0.03$ & $0.81 \pm 0.03$ & $0.99 \pm 0.00$ & $\mathbf{1.00 \pm 0.00}$ & $0.99 \pm 0.00$ & $0.98 \pm 0.01$ & $0.48 \pm 0.33$ \\ \cline{2-9}
    & \textbf{N-IID 0.1} & $0.28 \pm 0.01$ & $0.85 \pm 0.04$ & $0.91 \pm 0.07$ & $\mathbf{1.00 \pm 0.00}$ & $0.92 \pm 0.01$ & $0.92 \pm 0.01$ & $0.47 \pm 0.34$ \\ \hline
    \end{tabular}
    \caption{F1-scores of the proposed detection techniques for various settings with 1 FR client. Notations in the first column: A/V - AlexNet \& VGG-19, 10/100 - CIFAR10 \& CIFAR100, 4/8 clients, F/L - Fraboni \& Lin FR strategy. }
    \label{tab:ablation}
\end{table*}
\subsection{FR by Fraboni et al.}

~\cite{fraboni2021free} introduced a theoretical framework that formalizes FR attacks in FL schemes based on model averaging, relying on modelling SGD as a continuous time process. This framework provides theoretical guarantees on the convergence of the aggregated model to the optimal solution, which is critical for the attacker to not interfere with the learning process. They also introduced the \textit{disguised FR}, denoted as $\mathcal{F}^F$, which mimics SGD by adding a time-varying perturbation to the global model. Given a global model $M^{t-1}$, the FR client's update in round $t$ is defined in Eq.~\ref{eq:fraboni}, where $\alpha$ is a scaling factor, $\sigma$ is the standard deviation of the previous global gradient $G^{t-1}$ and $\gamma$ is the decay exponent, which ensures that as training progresses, the noise aligns with the natural reduction in SGD update magnitudes.

\begin{equation}
    \label{eq:fraboni}
    \mathcal{F}^F = M^{t-1} +\epsilon_{t} \hspace{0.5cm}
    \epsilon_t \sim \mathcal{N}(0, \alpha \cdot \sigma \cdot {t}^{-\gamma})
\end{equation}

\subsection{FR by Lin et al.}

~\cite{lin2019free} introduced the \textit{delta weights} attack, denoted as $\mathcal{F}^L$. This attack simulates a real update by adding noise to the global gradient of the previous round, as shown in Eq~\ref{eq:Lin}. Although the authors fixed the standard deviation $\sigma$ of the noise beforehand, we adapt the attack by computing the standard deviation similarly to~\cite{fraboni2021free}.

\begin{equation}
    \label{eq:Lin}
    \mathcal{F}^L = M^{t-1} + G^{t-1} + \mathcal{N}(0, \sigma_{F_L})
\end{equation}

\subsection{FR by Zhu et al.}

~\cite{zhu2021advanced} introduced the \textit{advanced} attack, denoted as $\mathcal{F}^Z$. This attack leverages geometrical properties inherent in IID settings to construct a FR update that is robust against feature-based detection defenses. Assuming that the optimal global model \( M^t \) converges to \( \sum_{n=1}^N \frac{|D_n|}{|D|} M_n^{t-1} \), the authors theoretically establish that the cosine relationship between honest client updates and the global model satisfies \( \mathbb{E}(\cos \beta) \approx C^2/(C^2 + e^{2\lambda t}) \) where \( C \) is a constant introduced to define the ratio between the expectations \( \mathbb{E}(\|M^t - M^0\|) = C \mathbb{E}(\|G_i^{t}\|) \). Furthermore, they derive the relationship for the \( \ell_2 \)-norm as follows.

\begin{equation*}
    \frac{\mathbb{E}(G_i)}{\mathbb{E}(\overline{G})} = \sqrt{\frac{n^2}{n + (n^2 - n) \mathbb{E}(\cos \beta)}}
\end{equation*}

Building on these properties, the attack injects Gaussian noise into the global gradient. Specifically, noise is added to a subset \( d \) of parameters from the previous round’s global gradient. The added noise is defined below where \( \|G_f^{t-1}\| \) is the \( \ell_2 \)-norm of the global gradient. This results in the final form of the free-rider update as presented in Equation~\ref{eq:Zhu}.

\begin{equation*}
    \varphi(t) = \sqrt{\frac{n^2}{n + (n^2 - n) \mathbb{E}(\cos \beta)} - 1} \|G_f^{t-1}\|_2
\end{equation*}

\begin{equation}
    \label{eq:Zhu}
    \mathcal{F}^Z = M^{t-1} + \frac{\|G^{t-1}\|_2}{\|G^{t-2}\|_2} G^{t-1} + \varphi(t) \mathcal{N}(0, d^{-1})
\end{equation}

\section{Membership Inference Attacks} \label{sec:appendix_mia}

\subsection{Yeom et al.}

~\cite{yeom2018privacy} characterizes the effect of overfitting on membership advantage, which is defined as how well an adversary can distinguish whether $x \sim S$ or $x \sim D$, where $S$ is the training set and $D$ the distribution from which the training set was drawn. They found a clear dependence of membership advantage on generalization error, leading them to propose a straightforward attack, which we call the \textit{Yeom} attack and noted as $\mathcal{M}^Y$. This attack is based on the prediction loss of a target sample $s$. The attacker infers membership if $\mathcal{L}(s,\theta)$ is lower than the expected training loss, given a subset of training samples that are part of the attacker's knowledge. 

\subsection{Li et al.}

~\cite{li2023effective} leverages the difference between the Gaussian distributions of cosine similarity for members and non-members of a client's training set to infer membership. For a global model $M^t(\theta)$, a client's gradient $G^t_n$, and a target sample ($x,y$), the cosine similarity is defined in Equation \ref{eq:cosine_attack} with the following inner product, given that $\tilde{G}_s^t = \nabla_{\theta}\mathcal{L}(y,x,M^t(\theta))$.
\begin{equation}
\label{eq:cosine_attack}
    CosSim(\tilde{G}_s^t, G_n^t) = \frac{\langle\tilde{G}_s^t, G_n^t\rangle}{(\|\tilde{G}_s^t\|_2\|G_n^t\|_2)}
\end{equation}

\section{Property Inference Attacks} \label{sec:appendix_pia}

We assume the label distributions are normalized, such that $\tilde{L^t_n} = L^t_n\cdot{(|L^t_n|}_1)^{-1}$.

\subsection{Wainakh et al.}

~\cite{wainakh2021user} relies on the gradients of the last layer where each node corresponds to one label. First, the impact of the labels are extracted by simulating training using $\tilde{D}$. For higher accuracy, an offset can also be estimated, which we left out, as it does not affect the FR detection. Then, the label distribution is extracted iteratively, by increasing the count of the label with the highest presence in the gradient and removing its impact. This is repeated until the number of labels in $|D_n|$ is extracted. 

\subsection{Dai et al.}

~\cite{dai2024decaf} is made up of three steps. First, the null classes (i.e., the labels that were not used during training) are removed. This is done by examining the gradient changes; the labels with no positive gradients are considered null classes. Then, the label basis (i.e., the gradient changes after training with a single type of label) and the unified base (i.e., the effect of training with the same amount of labels from all non-null classes) are extracted. This is done by simulating training for each of these scenarios using $\tilde{D}$. Finally, the label distribution for non-null classes is acquired by checking which linear combination of the basis (with non-negative coefficients) is the best to reproduce the gradients of the client. 

\end{document}